%% file: kPCA_PreImage.tex
\documentclass[lettersize,journal]{IEEEtran}
\usepackage{algorithmic}
\usepackage{algorithm}
\usepackage{array}
\usepackage[caption=false,font=footnotesize,labelfont=sf,textfont=sf]{subfig}
\usepackage{textcomp}
\usepackage{stfloats}
\usepackage{url}
\usepackage{verbatim}
\usepackage{graphicx}
\usepackage{cite}

\usepackage{amsmath, amssymb, amsthm, enumitem, mathtools, mathabx, graphicx, todonotes}
\usepackage[utf8]{inputenc} 
\usepackage[T1]{fontenc}    
\usepackage{url}            
\usepackage{booktabs}       
\usepackage{multirow}
\usepackage{amsfonts}       
\usepackage{nicefrac}       
\usepackage{microtype}      
\usepackage{xcolor}         
\usepackage{pgf}            

\usepackage[pdffitwindow=false,
            plainpages=false,
            pdfpagelabels=true,
            pdfpagemode=UseOutlines,
            pdfpagelayout=SinglePage,
            bookmarks=false,
            colorlinks=true,
            hyperfootnotes=false,
            linkcolor=blue,
            urlcolor=blue!30!black,
            citecolor=green!50!black]{hyperref}

\newtheorem{theorem}{Theorem}[section]

\newtheorem{definition}[theorem]{Definition}
\theoremstyle{definition}

\newtheorem{remark}[theorem]{Remark}
\newtheorem{assumption}[theorem]{Assumption}

\newcommand{\R}{\mathbb{R}}                                      
\renewcommand{\H}{\mathbb{H}}                                    
\newcommand{\I}{\mathrm{I}}                                      
\newcommand{\innerprod}[2]{\left\langle #1,\, #2 \right\rangle}  
\newcommand{\ts}{\hspace*{0.1em}}                                
\providecommand{\norm}[1]{\left\lVert #1 \right\rVert}           
\renewcommand{\vec}[1]{\boldsymbol{#1}}
\DeclareMathOperator{\mspan}{span}

\begin{document}

\title{Solving the pre-image problem in kernel PCA: \\ The reproducing property is all you need}

\author{Joel-Pascal Ntwali N'konzi, Stefan Klus
\thanks{J.P.N. N'konzi is with Maxwell Institute for Mathematical Sciences, University of Edinburgh \& Heriot--Watt University, Edinburgh, United Kingdom}
\thanks{S. Klus is with School of Mathematical and Computer Sciences, Heriot--Watt University, Edinburgh, United Kingdom}
}



\maketitle

\begin{abstract}
Kernel PCA extends PCA to nonlinear dimensionality reduction by implicitly mapping data into a reproducing kernel Hilbert space (RKHS) via the kernel trick, with applications ranging from image denoising to out-of-distribution detection. While its projection map is well understood, appropriately defining the associated reconstruction map is a longstanding challenge since projecting back from the RKHS to the input space is non-trivial. This is known as the \emph{pre-image problem}. Existing approaches have tackled this problem by either relying on nonlinear optimization, which is prone to numerical instabilities and sensitive to initialization, or by solving a separate supervised learning problem, incurring additional computational overhead. Furthermore, although the projection map of kernel PCA reduces to that of standard PCA when a linear kernel is used, most of these approaches are not guaranteed to recover PCA's reconstruction map in this case, a consistency property that one would naturally expect to hold. In this paper, we leverage the \emph{reproducing property} of the RKHS to derive a reconstruction map for kernel PCA that is provably consistent. Our derivations rely on the mild assumption that the RKHS either contains linear functions or the kernel is universal, which holds for standard kernels. We further extend the method to the random Fourier features (RFFs) approximation of kernel PCA by introducing a corresponding notion of \emph{RFF reproducing property}. Numerical experiments on synthetic and real-world datasets demonstrate that the proposed method performs comparably with state-of-the-art methods across kernel choices, while remaining more computationally efficient and maintaining consistent performance in settings where these baseline methods drastically fail.
\end{abstract}


\section{Introduction}

Principal component analysis (PCA)~\cite{pearson1901pca, hotelling1933analysis} is one of the most used tools from multivariate statistics with applications across all areas of science and engineering. PCA has deep ties with modern machine learning techniques. In particular, it has been shown that linear autoencoders (AEs) with the standard mean-squared error loss admit a global minimum given by the principal components (PCs)~\cite{baldi1989neural}. Furthermore, probabilistic PCA~\cite{tipping1999probabilistic, roweis1997algorithms} provides what is arguably the simplest case of a latent variable model and it has been used to investigate properties such as posterior collapse in nonlinear variational AEs~\cite{lucas2019don}. Despite its success, PCA is limited by the fact that it uncovers only linear relationships in the data. This prompted the development of nonlinear methods such as kernel PCA~\cite{mika1998kernel} and nonlinear AEs~\cite{hinton2006reducing}. The latter considers two functions $\vec{\varphi}_{\text{enc}} \colon \Omega \subseteq \R^d \to \R^q$ and $\vec{\varphi}_{\text{dec}}\colon \R^q \to \Omega$, called \emph{encoder} and \emph{decoder}, respectively, parameterized by neural networks such that $q \ll d$. Training is accomplished by minimizing the mean-squared reconstruction error, i.e., $\mathbb{E}_{\vec{x}} \norm{\vec{x} - \left(\vec{\varphi}_{\text{dec}} \circ \vec{\varphi}_{\text{enc}}\right)\left(\vec{x}\right)}_2^2$, where the expectation is estimated empirically based on training data. The objective function is nonlinear and typically non-convex, with no guarantee of attaining the global minimum. Kernel PCA (KPCA), on the other hand, leverages the well-known \emph{kernel trick} to first map the data into a (possibly infinite-dimensional) reproducing kernel Hilbert space (RKHS) $\H$~\cite{aronszajn1950theory} and then perform standard PCA in that space, yielding PCs that depend nonlinearly on the input data. Projecting data onto PCs requires kernel evaluations only, hence bypassing the need for non-convex optimization. Successful applications include image denoising~\cite{mika1998kernel, kim2005iterative,abrahamsen2009input,teixeira2008kpca,rasmussen2012nonlinear,bui2019projection}, feature extraction~\cite{rosipal2001kernel,jade2003feature,xu2007method,lopez2009svm,chang2015critical,peter20183d}, and out-of-distribution detection~\cite{samuel2016nonlinear,fang2024kernel}. Furthermore, although the exact formulation is computationally expensive in large-scale settings, approximations using random Fourier features (RFFs)~\cite{lopez2014randomized,ullah2018streaming,sriperumbudur2022approximate} and Nystr\"{o}m sampling~\cite{zhang2008improved,sterge2020gain,sterge2022statistical} are computationally efficient and have been shown to preserve statistical optimality under mild assumptions.

The challenging aspect in denoising and compression applications of KPCA is the fact that, in contrast to standard PCA, the decoder map does not trivially follow from the definition of the encoder. Since the projection is carried out in $\H$, one has to find a way of mapping back elements $\psi \in \H$ to corresponding points $\vec{x} \in \Omega$. This is known as the \emph{pre-image problem} in KPCA. Since an exact pre-image might not exist, early work proposed to use nonlinear optimization to find a point $\hat{\vec{x}}$ such that the squared distance between its image in the RKHS and $\psi$ is minimal~\cite{mika1998kernel, scholkopf1999input}. For isotropic kernels, such as the RBF kernel, this can be solved via a fixed-point iteration scheme. However, this approach is prone to numerical instability and sensitivity to initialization. Adding regularization to achieve numerical stability has been investigated, see, e.g., \cite{abrahamsen2009input, abrahamsen2011regularized}. To circumvent these issues, \cite{kwok2004pre} proposed a method based on calculating distances to a set of points both in the input and feature spaces, and to choose a pre-image point that satisfies structural constraints between the paired distances. This requires, in particular, that there is a one-to-one mapping between distances in the RKHS and the input space. This is the case for the RBF kernel and polynomial kernels of odd degree. In \cite{teixeira2008kpca}, the authors found that using the output from the method of \cite{kwok2004pre} to initialize the iterative scheme of \cite{mika1998kernel} performed better than each of the two methods separately. The authors of \cite{honeine2011closed} proposed to enforce isometry constraints and obtained an algorithm with a closed-form solution; see \cite{honeine2011preimage} for a review and \cite{honeine2026pre} for generalized iterative schemes.

A conceptually different approach was introduced in \cite{bakir2004learning}, where the authors proposed a supervised learning method to optimize a pre-image map from data. The idea is based on the fact that generating paired data is trivial: any point in the input space can be mapped to its corresponding RKHS element via the KPCA encoder map. Based on this observation, they used kernel ridge regression to learn a nonlinear map from the KPCA representation to the input space. The KPCA implementation contained in \emph{scikit-learn} uses this method for its inverse transform~\cite{scikit-learn}.  Despite being a major contribution, the method has some significant limitations. Firstly, in its fully flexible version, one has to specify two potentially different kernels: one for instantiating KPCA and the other for the inverse mapping. The canonical choice made by \emph{scikit-learn}'s implementation is to use a single kernel for both needs. Secondly, in large scale applications, one has to resort to approximations as exact kernel ridge regression becomes computationally infeasible. Thirdly, this approach does not recover standard PCA's decoder when a linear kernel is used. To the best of our knowledge, this consistency property does not hold for any of the existing pre-image methods, except that of \cite{mika1998kernel}. More recently, the authors of \cite{czaja2026comparative} proposed to use generative adversarial networks~\cite{goodfellow2014generative} to find pre-images in a denoising application. The main contributions of this paper are as follows:
\begin{enumerate}[leftmargin=3.5ex, itemsep=0ex, topsep=0.5ex, label=\roman*)]
	\item We leverage the structure of the associated RKHS, namely its \emph{reproducing property}, to show that KPCA admits a natural pre-image map under the assumption that it either contains linear functions or the kernel is universal, hence bypassing the need to solve a separate supervised learning problem. This assumption holds for kernels that are typically used in practice.
	\item We show that the proposed pre-image map is optimal in the least-squares sense, and that it coincides with PCA's reconstruction map for the linear kernel.
	\item By introducing a notion of \emph{RFF reproducing property}, we show that this method naturally extends to the RFF approximation of KPCA.
	\item Numerical experiments on synthetic and real-world datasets show that, across kernel choices, the proposed method performs either comparably with or significantly better than existing methods while being more computationally efficient.
\end{enumerate}

Section~\ref{sec:preliminaries} provides the necessary background. In Section~\ref{sec:kpca-preimage}, we introduce the proposed pre-image map for KPCA and show that it is both optimal and coincides with PCA's decoder for the linear kernel. We extend these results to the RFF approximation in Section~\ref{sec:kpca-rff-preimage}. Finally, we extensively evaluate the proposed method and compare it to state-of-the-art baselines using both synthetic and real-world datasets in Section~\ref{sec:experiments}. We conclude with a discussion in Section~\ref{sec:discussion}.

\section{Preliminaries}\label{sec:preliminaries}

\subsection{Standard PCA}
Consider training data of the form $\left\{\vec{x}^{(i)} \in \Omega \subseteq \R^d\right\}_{i=1}^m $. Let $\tilde{X} = \left[\tilde{\vec{x}}^{(1)}, \dots, \tilde{\vec{x}}^{(m)}\right]$ denote the centered data matrix, i.e.,  $\tilde{\vec{x}}^{(i)} = \vec{x}^{(i)} - \bar{\vec{x}} $, where $\bar{\vec{x}}$ is the mean of the data. Principal components found by PCA correspond to the normalized eigenvectors of the covariance matrix $C = \frac{1}{m} \tilde{X} \tilde{X}^\top$ with non-zero eigenvalues (note that all eigenvalues are nonnegative since $C$ is positive semi-definite). As is standard, we assume that the PCs are sorted in decreasing order of the associated eigenvalues. The first PC then captures the most variance in the data, and successive components capture the most of the remaining variance in directions orthogonal to the preceding ones. Denote the eigenpairs by $\left(\lambda_j, \vec{w}_j\right)_{j=1}^r$, where $r \leq d$ is the total number of PCs. Then any point $\vec{x} \in \Omega$ can be decomposed as
\begin{equation}\label{eq:pca-orthogonal-decomposition}
	\vec{x} = \bar{\vec{x}} + \sum_{j=1}^q \innerprod{\vec{x} - \bar{\vec{x}}}{\vec{w}_j}\vec{w}_j + \mathcal{P}_q^\perp\left(\vec{x} - \bar{\vec{x}}\right),
\end{equation}
where $q \le r$ is the number of retained components and $\mathcal{P}_q^\perp$ denotes the projection onto the orthogonal complement of $\mathcal{W}_q = \mspan\{\vec{w}_1, \dots, \vec{w}_q\}$. Therefore, PCA induces the projection map
\begin{equation}
	\vec{x} \mapsto \vec{z} = \vec{\varphi}_{\text{enc}}^\text{PCA}(\vec{x}; q) =  W_q^\top (\vec{x} - \bar{\vec{x}}) \in \R^q,
\end{equation}
with the associated reconstruction map
\begin{equation}
	\vec{z} \mapsto \hat{\vec{x}} = \vec{\varphi}_{\text{dec}}^\text{PCA}(\vec{z}; q) = \bar{\vec{x}} + W_q \vec{z} \approx \vec{x},
\end{equation}
where $W_q = \left[\vec{w}_1, \dots, \vec{w}_q \right] \in \R^{n \times q}$. This allows us to perform dimensionality reduction when $q \ll d$. It is worth noting that although $\vec{\varphi}_{\text{enc}}^\text{PCA} \circ \vec{\varphi}_{\text{dec}}^\text{PCA} = \vec{id}_{\R^q}$, the reverse does not hold in general, i.e., $\vec{\varphi}_{\text{dec}}^\text{PCA} \circ \vec{\varphi}_{\text{enc}}^\text{PCA} \neq \vec{id}_{\R^d}$. Nevertheless, $\vec{\varphi}_{\text{dec}}^\text{PCA}$ is the optimal affine reconstruction map in the least-squares sense. PCA is tightly connected to autoencoders in machine learning, see~\cite{baldi1989neural}.

\subsection{Kernel PCA}

Instead of performing PCA directly on the input data, KPCA~\cite{scholkopf1998nonlinear} leverages the kernel trick to first (implicitly) map the data into an RKHS and then performs PCA in this feature space.

Let $k\colon \Omega \times \Omega \to \R$ be a positive semi-definite kernel and $\H$ the associated RKHS~\cite{aronszajn1950theory}, with elements of $\H$ being real-valued functions on $\Omega$. Let $\vec{x} \mapsto \phi(\vec{x}) \in \H$ be the \emph{canonical feature map}, i.e., $\phi(\vec{x}) = k(\cdot, \vec{x})$. The defining characteristic of an RKHS is its \emph{reproducing property}, that is, $\forall f\in \H$, it holds that
\begin{equation}
	\innerprod{\phi(\vec{x})}{f}_{\H} = f(\vec{x}),
\end{equation}
see, e.g., \cite{scholkopf2002learning}. In particular, $\innerprod{\phi(\vec{x})}{\phi(\vec{y})}_{\H} = \phi(\vec{y})(\vec{x}) = k(\vec{x}, \vec{y})$ for all $\vec{x}, \vec{y} \in \Omega$. Hence, the kernel provides a way of evaluating inner products between featurized elements without explicitly mapping $\vec{x}$ and $\vec{y}$ into $\H$. To avoid cumbersome notation, we will omit the subscript $\H$ in what follows.

Let $\psi_1, \dots, \psi_s$ be the PCs in $\H$. Note that there can be up to $m$ PCs, i.e., $s \leq m$. The authors of \cite{scholkopf1998nonlinear} showed that
\begin{equation}
	\psi_j = \sum_{i=1}^{m} \alpha_{i,j} \tilde{\phi}\left(\vec{x}^{(i)}\right) = \vec{\alpha}_j^\top \vec{\tilde{\phi}}(X),\quad j=1, \dots, s,
\end{equation}
where $\vec{\tilde{\phi}}(X) = \left[\tilde{\phi}\left(\vec{x}^{(1)}\right), \dots, \tilde{\phi}\left(\vec{x}^{(m)}\right)\right]^\top \in \H^m$, and we are using the notation $\tilde{\phi}(\vec{x}) = \phi(\vec{x}) - \bar{\phi}$ with  $\bar{\phi} = \frac{1}{m} \sum_{i=1}^{m} \phi\left(\vec{x}^{(i)}\right)$ being the data mean in $\H$. Furthermore, the coefficient vectors $\vec{\alpha}_j = \left[\alpha_{1,j}, \dots, \alpha_{m,j}\right]^\top $  are eigenvectors of the \emph{centered} kernel matrix $\tilde{K}_{XX} = C_m K_{XX} C_m$ associated with non-zero eigenvalues. Here, $\left(K_{X X}\right)_{i,j} = k\left(\vec{x}^{(i)}, \vec{x}^{(j)}\right)$ and $C_m = \I - \frac{1}{m} \vec{1}_m \vec{1}_m^\top$ denotes the centering matrix, with $\vec{1}_m$ being the $m$-dimensional vector of ones. To ensure that the PCs have norm one in $\H$, each eigenvector is rescaled such that $\lambda_j \norm{\vec{\alpha}_j}_2^2 = 1$, where the $\lambda_j$ is the associated eigenvalue. In what follows, we assume this rescaling has been applied. Moreover, projection onto the PCs only requires kernel evaluations since, for $j=1, \dots, s$,
\begin{equation}\label{eq:kpca-projection}
	\innerprod{\tilde{\phi}(\vec{x})}{\psi_j} = \vec{\alpha}_j^\top \tilde{\vec{k}}_{\vec{x}},
\end{equation}
where $\tilde{\vec{k}}_{\vec{x}} = C_m \left[\vec{k}_{\vec{x}} - \frac{1}{m} K_{XX} \vec{1}_m \right]$ is the centered version of $\vec{k}_{\vec{x}} = \left[k\left(\vec{x}, \vec{x}^{(1)}\right), \dots, k\left(\vec{x}, \vec{x}^{(m)}\right)\right]^\top$.

\subsection{The pre-image problem}

Analogously to \eqref{eq:pca-orthogonal-decomposition}, we note that KPCA induces the following orthogonal decomposition in $\H$:
\begin{equation}\label{eq:kpca-orthogonal-decomposition}
	\tilde{\phi}(\vec{x}) = \sum_{j=1}^{q} \innerprod{\tilde{\phi}(\vec{x})}{\psi_j} \psi_j + \mathcal{P}_q^{\perp}\tilde{\phi}(\vec{x}).
\end{equation}
Then, by identifying $\{\psi_j\}_{j=1}^q$ with the canonical basis of $\R^q$, we obtain KPCA's encoder
\begin{equation}\label{eq:kpca-encoder}
	\vec{x} \mapsto \vec{z} = \vec{\varphi}_{\text{enc}}^{\text{KPCA}}(\vec{x}; q) = \begin{bmatrix}
		\innerprod{\tilde{\phi}(\vec{x})}{\psi_1} \\
		\vdots\\
		\innerprod{\tilde{\phi}(\vec{x})}{\psi_q}
	\end{bmatrix} = A_q^\top \tilde{\vec{k}}_{\vec{x}} \in \R^q,
\end{equation}
where $A_q = \left[\vec{\alpha}_1, \dots, \vec{\alpha}_q \right]\in \R^{m \times q}$. The challenging question remains how to define an appropriate decoder, i.e., a mapping $\vec{\varphi}_{\text{dec}}^{\text{KPCA}} \colon \R^q \to \Omega$ such that if $\vec{z} = \vec{\varphi}_{\text{enc}}^{\text{KPCA}}(\vec{x}, q)$, then $\vec{\varphi}_{\text{dec}}^{\text{KPCA}}(\vec{z}, q) \approx \vec{x}$. This is the longstanding \emph{pre-image problem} in kernel methods.

\section{A reproducing property solution to the KPCA pre-image problem}\label{sec:kpca-preimage}

We show that the pre-image problem in KPCA admits a natural solution. We achieve this by leveraging the reproducing property to show that KPCA leads to an input-space decomposition akin to \eqref{eq:pca-orthogonal-decomposition}. Our derivations rely on either of the following two assumptions being true.

\begin{assumption}\label{assumption-main}
	Let $\pi_i\colon \Omega \to \R$ be the projection map onto the $i^{th}$ component, i.e., $\pi_i(\vec{x}) = x_i$ for all $\vec{x} \in \Omega$. Then $\pi_i \in \H$ for $i = 1, \dots, d$.
\end{assumption}

\begin{assumption}\label{assumption-main-universality}
	The input space $\Omega$ is compact and the kernel $k$ is \emph{universal} in the sense of \cite{steinwart2001influence}; i.e., it is continuous and the induced RKHS $\H$ is dense in the space $\mathcal{C}(\Omega)$ of continuous functions on $\Omega$ with respect to the uniform norm.
\end{assumption}

We start by presenting our main results and discuss the scope of their validity in Section~\ref{sec:scope-of-results}.

\begin{remark}\label{remark:continuous-extension}
	When working with Assumption~\ref{assumption-main-universality}, we extend $\innerprod{f}{\cdot}$ from $\H$ to $\mathcal{C}(\Omega)$ by continuity and use the same symbol for this extension; i.e., whenever $\left(g_{\ell}\right)_{\ell=0}^\infty$ converges to $g \in \mathcal{C}(\Omega)$, with $g_{\ell} \in \H$ for all $\ell$, then
	\begin{equation}
		\innerprod{f}{g} \coloneqq \lim_{\ell \to \infty} \innerprod{f}{g_{\ell}},
	\end{equation}
    for any $f \in \H$.
\end{remark}

\subsection{Main results}
Our derivations rely on the evaluation of inner products of the form $\innerprod{\psi}{\pi_i}$, where $\psi \in \H$. In what follows, we use $\innerprod{\psi}{\vec{\pi}}$ to denote the component-wise inner product between $\psi$ and $\vec{\pi} = \left[\pi_1, \dots, \pi_d\right]^\top$. When such evaluations are justified, the reproducing property implies that
\begin{equation}\label{eq:key-equality}
	\innerprod{\phi(\vec{x})}{\vec{\pi}}  = \vec{\pi}(\vec{x}) = \vec{x}.
\end{equation}
This equality is the basis of all our results. In particular, we obtain the following proposition.

\begin{theorem}\label{thm:injectivity}
	Under Assumption \ref{assumption-main}, the  mapping $\R^d \ni \vec{x} \mapsto \phi(\vec{x}) \in \H$ is injective. This also holds for the centered version $ \vec{x} \mapsto \tilde{\phi}(\vec{x}) = \phi(\vec{x}) - \bar{\phi}$.
\end{theorem}
\begin{proof}
	Let $\vec{x}, \vec{y} \in \R^d$ and assume that $\phi(\vec{x}) = \phi(\vec{y})$. Then $\vec{x} = \innerprod{\phi(\vec{x})}{\vec{\pi}} = \innerprod{\phi(\vec{y})}{\vec{\pi}} = \vec{y}$, showing that the map $\vec{x} \mapsto \phi(\vec{x})$ is injective. The second part of the proposition follows from the fact that $\tilde{\phi}(\vec{x}) = \tilde{\phi}(\vec{y})$ implies $\phi(\vec{x}) = \phi(\vec{y})$, and vice versa.
\end{proof}

The implication is that if $\psi \in \H$ admits an exact pre-image~$\hat{\vec{x}}$, then $\hat{\vec{x}}$ is unique. We now state our main result.

\begin{theorem}[\textbf{KPCA reconstruction map}]
	If either of Assumption \ref{assumption-main} or Assumption~\ref{assumption-main-universality} holds, then the decomposition
	\begin{equation}
		\tilde{\vec{x}} = \sum_{j=1}^{q} \innerprod{\tilde{\phi}(\vec{x})}{\psi_j} \tilde{X} \vec{\alpha}_j + \check{\vec{r}}(\vec{x}) = \tilde{X} A_q \vec{\varphi}_{\text{enc}}^{\text{KPCA}}(\vec{x}; q) + \check{\vec{r}}(\vec{x}),
	\end{equation}
	with $\check{\vec{r}}(\vec{x}) = \innerprod{\mathcal{P}_q^{\perp}\tilde{\phi}(\vec{x})}{\vec{\pi}}$ also holds. This induces the following decoder map
	\begin{equation}\label{eq:kpca-preimage-map}
		\R^q \ni \vec{z} \mapsto \vec{\varphi}_{\text{dec}}^{\text{KPCA}}(\vec{z}; q) = \bar{\vec{x}} + \tilde{X} A_q \vec{z}.
	\end{equation}
	In particular, $\vec{\varphi}_{\text{dec}}^{\text{KPCA}}$ leads to an exact reconstruction of the training data when $q = s$.
\end{theorem}
\begin{proof}
	Let $\vec{x} \in \Omega$.
	\paragraph{Assume Assumption~\ref{assumption-main} holds} It follows from \eqref{eq:kpca-orthogonal-decomposition} that
		\begin{equation}\label{eq:inner-prod-phi-tilde-pi}
			\innerprod{\tilde{\phi}(\vec{x})}{\vec{\pi}} = \sum_{j=1}^{q} \innerprod{\tilde{\phi}(\vec{x})}{\psi_j} \innerprod{\psi_j}{\vec{\pi}}  + \innerprod{\mathcal{P}_q^{\perp}\tilde{\phi}(\vec{x})}{\vec{\pi}}.
		\end{equation}
		However,
		\begin{align*}
			\innerprod{\tilde{\phi}(\vec{x})}{\vec{\pi}} &= \innerprod{\phi(\vec{x})}{\vec{\pi}} -  \innerprod{\bar{\phi}}{\vec{\pi}}\\
			&=  \vec{x} -  \frac{1}{m} \sum_{i=1}^{m} \innerprod{\phi\left(\vec{x}^{(i)}\right)}{\vec{\pi}} \tag*{\text{(by \eqref{eq:key-equality})}}\\
			&= \vec{x} - \bar{\vec{x}} = \tilde{\vec{x}},
		\end{align*}
		and
        \begin{align*}
            \innerprod{\psi_j}{\vec{\pi}} &= \Big\langle \sum_{i=1}^{m} \tilde{\alpha}_{ij} \tilde{\phi}\left(\vec{x}^{(i)}\right), \vec{\pi} \Big\rangle
			 = \sum_{i=1}^{m} \tilde{\alpha}_{ij} \innerprod{\tilde{\phi}\left(\vec{x}^{(i)}\right)}{\vec{\pi}} \\
			 &= \sum_{i=1}^{m} \tilde{\alpha}_{ij} \tilde{\vec{x}}^{(i)} = \tilde{X} \vec{\alpha}_j.
        \end{align*}
        Hence, \eqref{eq:inner-prod-phi-tilde-pi} is equivalent to
		\begin{align*}
			\tilde{\vec{x}} &= \sum_{j=1}^{q} \innerprod{\tilde{\phi}(\vec{x})}{\psi_j} \tilde{X} \vec{\alpha}_j + \innerprod{\mathcal{P}_q^{\perp}\tilde{\phi}(\vec{x})}{\vec{\pi}} \\
			&= \tilde{X} A_q \vec{\varphi}_{\text{enc}}^{\text{KPCA}}(\vec{x}; q) + \innerprod{\mathcal{P}_q^{\perp}\tilde{\phi}(\vec{x})}{\vec{\pi}}.
		\end{align*}

	\paragraph{Assume Assumption~\ref{assumption-main-universality} holds} Since $\pi_i \in \mathcal{C}(\Omega)$, it follows from the density of $\H$ that there exists a sequence $\left(p_{i,\ell}\right)_{\ell=0}^\infty \subset \H$ such that $\pi_i = \lim_{\ell \to \infty} p_{i,\ell}$ uniformly, for $i=1, \dots, d$. Denote $\vec{p}_{\ell} = \left[p_{1,\ell}, \dots, p_{d, \ell}\right]^\top$ for all $\ell$. It follows from \eqref{eq:kpca-orthogonal-decomposition} that
		\begin{multline*}
  \lim_{\ell \to \infty} \innerprod{\tilde{\phi}(\vec{x})}{\vec{p}_{\ell}} = \sum_{j=1}^{q}  \lim_{\ell \to \infty} \innerprod{\tilde{\phi}(\vec{x})}{\psi_j} \innerprod{\psi_j}{\vec{p}_{\ell}} \\
  +  \lim_{\ell \to \infty} \innerprod{\mathcal{P}_q^{\perp}\tilde{\phi}(\vec{x})}{\vec{p}_{\ell}}.
		\end{multline*}
 Moreover,
		\begin{align*}
			\lim_{\ell \to \infty} \innerprod{\tilde{\phi}(\vec{x})}{\vec{p}_{\ell}} &= \lim_{\ell \to \infty} \innerprod{\phi(\vec{x})}{\vec{p}_{\ell}} \\
			&\qquad - \frac{1}{m} \sum_{i=1}^m \lim_{\ell \to \infty} \innerprod{\phi\left(\vec{x}^{(i)}\right)}{\vec{p}_{\ell}}\\
			&= \lim_{\ell \to \infty} \vec{p}_{\ell}(\vec{x}) - \frac{1}{m} \sum_{i=1}^m \lim_{\ell \to \infty} \vec{p}_{\ell}\left(\vec{x}^{(i)}\right) \\
			&= \vec{\pi}(\vec{x}) - \frac{1}{m} \sum_{i=1}^m \vec{\pi}\left(\vec{x}^{(i)}\right)\\
			&= \vec{x} - \bar{\vec{x}} = \tilde{\vec{x}},
		\end{align*}
    where we used the the reproducing property and the uniform to point-wise convergence. Analogously, it holds that $\lim_{\ell \to \infty} \innerprod{\psi_j}{\vec{p}_{\ell}} = \tilde{X} \vec{\alpha}_j$ and $\lim_{\ell \to \infty} \innerprod{\mathcal{P}_q^{\perp}\tilde{\phi}(\vec{x})}{\vec{p}_{\ell}} = \innerprod{\mathcal{P}_q^{\perp}\tilde{\phi}(\vec{x})}{\vec{\pi}}$ by Remark~\ref{remark:continuous-extension}.
\end{proof}

The following result shows that the derived KPCA pre-image map is consistent with standard PCA.
\begin{theorem}\label{cor:preimage-consistency}
	For the linear kernel $k(\vec{x}, \vec{x}') = \vec{x}^\top \vec{x}'$ and number of PCs $q \leq d$, the KPCA pre-image map in \eqref{eq:kpca-preimage-map} coincides with standard PCA's decoder.
\end{theorem}

\begin{proof}
	It suffices to show that $\tilde{X} A_q = W_q$, where $W_q$ contains the first $q$ normalized eigenvectors of the covariance matrix $C =\frac{1}{m} \tilde{X} \tilde{X}^\top$. For the linear kernel, $\tilde{K} = \tilde{X}^\top \tilde{X}$. Let $\tilde{K} = U \Lambda U^\top$ be the eigendecomposition of $\tilde{K}$ and $\tilde{X} = R \Sigma Q^\top$ the singular-value decomposition (SVD) of $\tilde{X}$. It follows that $\tilde{K} = \tilde{X}^\top \tilde{X} = Q \Sigma^2 Q^\top$, i.e., $Q = U$ and $\Sigma^2 = \Lambda$. Also, by construction, $A_q = U_q \Lambda_q^{-1/2}$. Hence,
	\begin{equation*}
		\tilde{X} A_q = \tilde{X} Q_q \Sigma_q^{-1} = \underbrace{R \Sigma Q^\top Q_q}_{R_q \Sigma_q} \Sigma_q^{-1} = R_q.
	\end{equation*}
	On the other hand, $C = \frac{1}{m} \tilde{X} \tilde{X}^\top = \frac{1}{m} R \Sigma^2 R^\top$. That is, $R$ is the matrix of normalized eigenvectors of $C$. In particular, $R_q = W_q$.
\end{proof}

The derived decoder is related to supervised learning-based pre-image methods in the following way:
\begin{theorem}\label{thm:kpca-preimage-optimal}
	 $\vec{\varphi}_{\text{dec}}^{\text{KPCA}}$ is the optimal affine reconstruction map in the least-squares sense.
\end{theorem}

\begin{proof}
	The solution to the least-squares problem
	\begin{align*}
		\hat{W} &= \underset{W \in \R^{d\times q}}{\arg\min} \frac{1}{m} \sum_{i=1}^m \norm{\tilde{\vec{x}}^{(i)} - W \vec{\varphi}_{\text{enc}}^{\text{KPCA}}\left(\vec{x}^{(i)}; q\right) }_2^2 \\
		&= \underset{W \in \R^{d\times q}}{\arg\min} \frac{1}{m} \norm{\tilde{X} - W Z }_F^2,
	\end{align*}
	is given by $\hat{W} = \tilde{X} Z^\dagger$. Here,
	\begin{equation*}
    Z =
	\begin{bmatrix}
		\vec{\varphi}_{\text{enc}}^{\text{KPCA}}\left(\vec{x}^{(1)}; q\right) &
		\dots &
		\vec{\varphi}_{\text{enc}}^{\text{KPCA}}\left(\vec{x}^{(m)}; q\right)
	\end{bmatrix}
    \end{equation*}
	and $Z^\dagger$ denotes its Moore--Penrose pseudoinverse. In addition, we have from \eqref{eq:kpca-encoder} that $Z = A_q^\top \tilde{K}_{XX}$. Let $\tilde{K}_{XX} = U \Lambda U^\top$ be the SVD of $\tilde{K}_{XX}$. Note that since $\tilde{K}_{XX}$ is symmetric positive semi-definite, its eigenvalues and singular values coincide. Also, by definition, $A_q = U_q \Lambda^{-1/2}$. It follows that
	\begin{align*}
		Z &= A_q^\top \tilde{K}_{XX} =  \left(\Lambda_q^{-1/2} U_q^\top\right) \left(U \Lambda U^\top\right)\\
		&= \Lambda_q^{-1/2} \Lambda_q U_q^\top = \Lambda_q^{1/2}  U_q^\top,
	\end{align*}
	and
	$
	\hat{W} = \tilde{X} Z^\dagger = \tilde{X} (\Lambda_q^{1/2}  U_q^\top)^\dagger = \tilde{X} U_q \Lambda_q^{-1/2} = \tilde{X} A_q
	$.
\end{proof}

It is important to note that this result does not rely on Assumptions \ref{assumption-main} or \ref{assumption-main-universality}. Its proof then provides an alternative view of $\vec{\varphi}_{\text{dec}}^{\text{KPCA}}$ as the solution of a minimization problem over the class of reconstruction functions of the form $\vec{z} \mapsto \bar{\vec{x}} + W \vec{z}$. This perspective is insightful and allows us, in particular, to obtain the \emph{Tikhonov regularized pre-image map}
\begin{equation}
	\vec{z} \mapsto \vec{\varphi}_{\text{dec}}^{\text{KPCA}}(\vec{z}; q, \delta) = \bar{\vec{x}} + \tilde{X} A_q \Lambda_q (\Lambda_q + m\delta I)^{-1} \vec{z},
\end{equation}
where $\delta >0 $ is the regularization parameter.

\subsection{Scope of our results}\label{sec:scope-of-results}
Our results rely on either of Assumptions~\ref{assumption-main} or \ref{assumption-main-universality} being true. This has a broad scope of validity in practice for the following reasons:
	\begin{itemize}
		\item Assumption~\ref{assumption-main} holds for the linear and polynomial kernels on any domain, and for general \emph{dot-product kernels} of the form $k(\vec{x}, \vec{y}) = \zeta(\vec{x}^\top \vec{y})$ with $\zeta(t) = \sum_{n=0} a_n t^n$ being analytic with non-negative coefficients, i.e., $a_n \geq 0$ for all $n$. This includes the exponential dot-product kernel. Moreover, such kernels are universal~\cite{steinwart2001influence,micchelli2006universal}. It also holds for sufficiently smooth Mat\'{e}rn kernels based on the fact that the corresponding RKHSs are norm-equivalent to Sobolev spaces, see, e.g., \cite{wendland2005scattered, kanagawa2018gaussian}.
		\item In practical applications, the data is contained in a compact subdomain of $\R^d$. On such domains, popular kernels such as the Gaussian RBF, Laplace, Mat\'{e}rn, and rational quadratic kernels are universal~\cite{steinwart2001influence}, and hence Assumption~\ref{assumption-main-universality} is valid; see, e.g., \cite{steinwart2001influence, micchelli2006universal,sriperumbudur2011universality} for more details on universal kernels.
		\item It follows from RKHS properties~\cite{aronszajn1950theory} that once a base kernel that satisfies either of the two assumptions has been identified, new kernels that satisfy the same assumption can be easily constructed using the sum of the base kernel and any other positive-definite kernel.
	\end{itemize}

\subsection{Relationship to optimization-based methods}

Earlier works defined an approximate pre-image for an input point $\vec{x} \in \Omega$ as the solution to the optimization problem
	\begin{equation}\label{eq:preimage-optimization}
		\vec{\hat{y}} = \underset{\vec{y} \in \Omega}{\arg\min} \norm{\tilde{\phi}(\vec{y}) - f_{\vec{x}}}^2,
	\end{equation}
where $f_{\vec{x}} = \mathcal{P}_q\tilde{\phi}(\vec{x}) \in \H$ is the KPCA representation of $\vec{x}$. In general, this problem is nonlinear and non-convex, and needs to be solved numerically with no guarantee of obtaining a global minimizer~\cite{mika1998kernel,scholkopf1999input,abrahamsen2009input}. Moreover, these methods do not come with theoretical guarantees of finding the exact pre-image when it exists, i.e., a point $\vec{\hat{y}} \in \Omega$ such that $\tilde{\phi}(\vec{\hat{y}}) = f_{\vec{x}}$ . In contrast, if such a point exists, it would correspond exactly to the output from our method since it can be found via the inner product evaluation, i.e.,
	\begin{equation*}
		\vec{\hat{y}} - \vec{\bar{x}} = \vec{\tilde{\hat{y}}} = \innerprod{\tilde{\phi}(\vec{\hat{y}})}{\vec{\pi}} = \innerprod{f_{\vec{x}}}{\vec{\pi}}.
	\end{equation*}

Furthermore, in the more common setting where an exact pre-image does not exist, the squared error between the output from the derived KPCA decoder and the solution of \eqref{eq:preimage-optimization} is bounded. To see that, let $\hat{\epsilon}_{\vec{x}} = \tilde{\phi}(\vec{\hat{y}}) - f_{\vec{x}} \in \H$ be the minimal error vector. Then
	\begin{align*}
		\vec{\tilde{\hat{y}}} &= \innerprod{\tilde{\phi}(\vec{\hat{y}})}{\vec{\pi}} = \innerprod{f_{\vec{x}}}{\vec{\pi}} + \innerprod{\hat{\epsilon}_{\vec{x}} }{\vec{\pi}} = \vec{\tilde{\hat{x}}} + \innerprod{\hat{\epsilon}_{\vec{x}} }{\vec{\pi}},
	\end{align*}
where $\vec{\tilde{\hat{x}}} = \tilde{X} A_q \vec{\varphi}_{\text{enc}}^{\text{KPCA}}(\vec{x}; q)$ and $\vec{\hat{x}} = \vec{\bar{x}} + \vec{\tilde{\hat{x}}}$ is the output from the derived KCPA decoder. It follows that
	\begin{align*}
		\norm{\vec{\hat{y}} - \vec{\hat{x}}}^2 &= \norm{\innerprod{\hat{\epsilon}_{\vec{x}} }{\vec{\pi}}}^2 = \sum_{i=1}^d \innerprod{\hat{\epsilon}_{\vec{x}} }{\pi_i}^2 \\
		& \leq \sum_{i=1}^d \norm{\epsilon_{\vec{x}}}^2 \norm{\pi_i}^2 = \norm{\epsilon_{\vec{x}}}^2 \norm{\vec{\pi}}^2,
	\end{align*}
using the Cauchy--Schwarz inequality. Additionally, since $\vec{\hat{y}}$ is optimal, the squared error $\norm{\epsilon_{\vec{x}}}^2$ factor is bounded from above by the KPCA projection error, i.e., $\norm{\epsilon_{\vec{x}}}^2 \leq \norm{\tilde{\phi}(\vec{x}) - f_{\vec{x}}}^2 = \norm{\mathcal{P}_q^\perp \tilde{\phi}(\vec{x})}^2$, which, on training data points, decays to zero as $q$ tends to the rank of the kernel matrix. In practical applications, this decay will be sharp since kernel matrices typically have a spectral gap. In summary, we obtain the following error bounds
	\begin{equation}
		\norm{\vec{\hat{y}} - \vec{\hat{x}}}^2 \leq \norm{\epsilon_{\vec{x}}}^2 \norm{\vec{\pi}}^2 \leq \norm{\mathcal{P}_q^\perp \tilde{\phi}(\vec{x})}^2 \norm{\vec{\pi}}^2,
	\end{equation}
which are agnostic to the structure of the used kernel. To the best of our knowledge, such theoretical guarantees are not available for any of the other methods.

\section{Extension to randomized principal component analysis}
\label{sec:kpca-rff-preimage}

As typical of kernel methods, KPCA scales cubicly with the training data size. To make KPCA applicable in large data settings, the authors of \cite{lopez2014randomized} proposed to use random Fourier features (RFFs) to approximate kernel evaluations and called the resulting method \emph{randomized principal component analysis}. This approach is based on the seminal work of Rahimi and Recht \cite{rahimi2007random} and requires the kernel to be translation-invariant. Due to the connection with KPCA, we refer to this method as KPCA-RFF.

Let $\vec{x} \mapsto \vec{\xi}(\vec{x}) \in \R^{n} $ denote the mapping onto random features such that
\begin{equation}\label{eq:rff-approx}
	k(\vec{x}, \vec{x}') \approx \frac{1}{n} \sum_{i=1}^n \xi_i(\vec{x}) \xi_i(\vec{x}') = \frac{1}{n} \innerprod{\vec{\xi}(\vec{x})}{\vec{\xi}(\vec{x}')}_{\R^n},
\end{equation}
where $n$ is the number of random projections. KPCA-RFF works by first passing the input data through $\vec{\xi}$ and then applying standard PCA. This induces the decomposition
\begin{equation}\label{eq:rff-kpca-orthogonal-decomposition}
	\tilde{\vec{\xi}}(\vec{x}) = \sum_{j=1}^q \innerprod{\tilde{\vec{\xi}}(\vec{x})}{\vec{v}_j}_{\R^n} \vec{v}_j + \mathcal{P}_q^{\perp}\tilde{\vec{\xi}}(\vec{x}),
\end{equation}
where the PCs $\vec{v}_j$ are given by the normalized eigenvectors of the covariance matrix $C_{\vec{\xi}} = \frac{1}{m} \tilde{\Xi}_X \tilde{\Xi}_X^\top$. Here, $\tilde{\Xi}_X = \begin{bmatrix}
	\tilde{\vec{\xi}}\left(\vec{x}^{(1)}\right) & \dots & \tilde{\vec{\xi}}\left(\vec{x}^{(m)}\right)
\end{bmatrix} $ is the centered data matrix in $\R^n$. To extend the results from Section \ref{sec:kpca-preimage} to KPCA-RFF, we introduce the (approximate) reproducing property for RFF-based methods.

\begin{definition}
	Let $f \in \H$ and consider its representation in terms of the canonical feature map
	\begin{equation}\label{eq:rkhs-feature-representation}
		f = \sum_{i=1}^{\infty} c_{i, f} \phi\left(\vec{x}^{(i)}_f \right),
	\end{equation}
for some constants $c_{i, f}$ and vectors $\vec{x}^{(i)}_f$ in $\R^d$ such that the series converges. We define $\mathcal{P}_{\vec{\xi}} \colon \H \to \R^n$ as the map
	\begin{equation}
		f = \sum_{i=1}^{\infty} c_{i, f} \phi\left(\vec{x}^{(i)}_f \right) \longmapsto \mathcal{P}_{\vec{\xi}}(f) =  \sum_{i=1}^{\infty} n^{-\nicefrac{1}{2}} c_{i, f} \vec{\xi} \left(\vec{x}^{(i)}_f \right).
	\end{equation}
\end{definition}

Consider $\vec{x} \in \R^d$, then
\begin{align*}
	\innerprod{n^{-\nicefrac{1}{2}} \vec{\xi}(\vec{x})}{\mathcal{P}_{\vec{\xi}}(f)}_{\R^n} &= \innerprod{n^{-1/2} \vec{\xi}(\vec{x})}{\sum_{i=1}^{\infty} n^{-\nicefrac{1}{2}} c_{i, f} \vec{\xi} \left(\vec{x}^{(i)}_f \right)}_{\R^n}\\
	&=  \sum_{i=1}^{\infty} n^{-1} c_{i, f} \innerprod{ \vec{\xi}(\vec{x})}{\vec{\xi}\left(\vec{x}^{(i)}_f \right)}_{\R^n} \\
	&\approx \sum_{i=1}^{\infty} c_{i, f} \innerprod{ \phi(\vec{x})}{\phi\left(\vec{x}^{(i)}_f\right)}_{\H}    \tag*{\text{(by definition of RFFs in \eqref{eq:rff-approx})}}\\
	&= \innerprod{ \phi(\vec{x})}{\sum_{i=1}^{\infty} c_{i, f} \phi\left(\vec{x}^{(i)}_f\right)}_{\H}\\
	&= \innerprod{ \phi(\vec{x})}{f}_{\H}   \tag*{\text{(by representation of $f$ in \eqref{eq:rkhs-feature-representation})}}\\
	&= f(\vec{x}) \tag*{\text{(by the reproducing property)}}.
\end{align*}

\begin{theorem}[\textbf{Reproducing property for RFFs}]\label{thm:rff-reproducing-property}
Given the random Fourier feature approximation \eqref{eq:rff-approx} of a shift-invariant positive semi-definite kernel $k(\cdot, \cdot)$, the following approximate reproducing property holds: For any $f \in \H$, we have
	\begin{equation}
		\innerprod{n^{-1/2} \vec{\xi}(\vec{x})}{\mathcal{P}_{\vec{\xi}}(f)}_{\R^n} \approx f(\vec{x}),
	\end{equation}
	for any $f \in \H$.
\end{theorem}

Intuitively,  ``$\approx$'' can be replaced by an equality as $n$ tends to infinity. Deriving proper convergence rates is, however, beyond the scope of this paper; see, e.g., \cite{rahimi2007random, sterge2020gain, sterge2022statistical, rudi2017generalization}.  We leverage this (approximate) reproducing property to obtain a decomposition of points in the input space in terms of PCs obtained via KPCA-RFF.

\begin{theorem}[\textbf{KPCA-RFF reconstruction map}]\label{thm:rff-preimage-map}
	If the kernel is translation-invariant and either of Assumptions \ref{assumption-main} or \ref{assumption-main-universality} holds, the following \underline{approximate} decomposition
	\begin{align*}
		\tilde{\vec{x}} &\approx \sum_{j=1}^{q} \frac{1}{m \lambda_j} \innerprod{\tilde{\vec{\xi}}(\vec{x})}{\vec{v}_j} \tilde{X} \tilde{\Xi}_X^\top \vec{v}_j + \check{\vec{r}}(\vec{x}) \\
		&= \frac{1}{m} \tilde{X} \tilde{\Xi}_X^\top V_q \Lambda_q^{-1} \vec{\varphi}_{\text{enc}}^{\text{KPCA-RFF}}(\vec{x}; q) + \breve{\vec{r}}(\vec{x}),
	\end{align*}
	with
	$
	\vec{x} \mapsto \vec{\varphi}_{\text{enc}}^{\text{KPCA-RFF}}(\vec{x}; q) = V_q^\top \tilde{\vec{\xi}}(\vec{x}) \in \R^q
	$
	and
	$
	\breve{\vec{r}}(\vec{x}) = n^{-1/2} \innerprod{ \mathcal{P}_q^{\perp}\tilde{\vec{\xi}}(\vec{x})}{\mathcal{P}_{\vec{\xi}}(\vec{\pi}) }
	$ holds. This induces the decoder map
	\begin{equation*}
		\R^q \ni \vec{z} \mapsto \vec{\varphi}_{\text{dec}}^{\text{KPCA-RFF}}(\vec{z}; q) = \bar{\vec{x}} + \frac{1}{m} \tilde{X} \tilde{\Xi}_X^\top V_q \Lambda_q^{-1} \vec{z}.
	\end{equation*}
\end{theorem}

\begin{proof}
	As before, we start with the decomposition \eqref{eq:rff-kpca-orthogonal-decomposition}. Multiplying both sides of the equation by $n^{-1/2}$ and calculating the inner product with $\mathcal{P}_{\vec{\xi}}(\vec{\pi})$, we obtain
	\begin{multline}\label{eq:inner-prod-rff-preimage-proof}
		\innerprod{n^{-\nicefrac{1}{2}} \tilde{\vec{\xi}}(\vec{x})}{\mathcal{P}_{\vec{\xi}}(\vec{\pi})} = \sum_{j=1}^q \innerprod{\tilde{\vec{\xi}}(\vec{x})}{\vec{v}_j} \innerprod{n^{-\nicefrac{1}{2}}\vec{v}_j}{\mathcal{P}_{\vec{\xi}}(\vec{\pi})} \\
		+ \breve{\vec{r}}(\vec{x}).
	\end{multline}
	On one hand, it follows from Theorem \ref{thm:rff-reproducing-property} that
	\begin{align*}
		\innerprod{n^{-1/2} \tilde{\vec{\xi}}(\vec{x})}{\mathcal{P}_{\vec{\xi}}(\vec{\pi})} &= \innerprod{n^{-1/2} \vec{\xi}(\vec{x})}{\mathcal{P}_{\vec{\xi}}(\vec{\pi})} \\
		&- \frac{1}{m} \sum_{i=1}^m \innerprod{n^{-1/2}  \vec{\xi}\left(\vec{x}^{(i)}\right)}{\mathcal{P}_{\vec{\xi}}(\vec{\pi})} \\
		&\approx \vec{\pi}(\vec{x}) - \frac{1}{m} \sum_{i=1}^m  \vec{\pi}\left(\vec{x}^{(i)}\right) \\
		&= \vec{x} - \frac{1}{m} \sum_{i=1}^m \vec{x}^{(i)} = \tilde{\vec{x}}.
	\end{align*}
	On the other hand, since the $\vec{v}_j$'s are eigenvectors of the covariance matrix $C_{\vec{\xi}}$, we may write $\vec{v}_j = \lambda_j^{-1} C_{\vec{\xi}} \vec{v}_j $, where $\lambda_j$ is the associated eigenvalue. Hence,
	\begin{align*}
		&\innerprod{n^{-1/2} \vec{v}_j}{\mathcal{P}_{\vec{\xi}}(\vec{\pi})} =  \lambda_j^{-1} \innerprod{n^{-1/2} C_{\vec{\xi}} \vec{v}_j}{\mathcal{P}_{\vec{\xi}}(\vec{\pi})} \\
		&= \frac{1}{m \lambda_j} \innerprod{n^{-1/2} \sum_{i=1}^m \tilde{\vec{\xi}} \left(\vec{x}^{(i)}\right) \tilde{\vec{\xi}}^\top\left(\vec{x}^{(i)}\right) \vec{v}_j}{\mathcal{P}_{\vec{\xi}}(\vec{\pi})}\\
		&= \frac{1}{m \lambda_j} \sum_{i=1}^m  \innerprod{\tilde{\vec{\xi}}\left(\vec{x}^{(i)}\right)}{\vec{v}_j} \innerprod{n^{-1/2} \tilde{\vec{\xi}}\left(\vec{x}^{(i)}\right)}{\mathcal{P}_{\vec{\xi}}(\vec{\pi})}\\
		&\approx \frac{1}{m \lambda_j} \sum_{i=1}^m  \innerprod{\tilde{\vec{\xi}}\left(\vec{x}^{(i)}\right)}{\vec{v}_j} \tilde{\vec{x}}^{(i)} = \frac{1}{m \lambda_j} \tilde{X} \tilde{\Xi}_X^\top \vec{v}_j.
	\end{align*}
	Putting everything together, we obtain from \eqref{eq:inner-prod-rff-preimage-proof} that
	\begin{align*}
		\tilde{\vec{x}} &\approx \sum_{j=1}^q \frac{1}{m \lambda_j} \innerprod{\tilde{\vec{\xi}}(\vec{x})}{\vec{v}_j} \tilde{X} \tilde{\Xi}_X^\top \vec{v}_j + \breve{\vec{r}}(\vec{x}) \\
		&= \frac{1}{m} \tilde{X} \tilde{\Xi}_X^\top V_q \Lambda_q^{-1} \left(V_q^\top \tilde{\vec{\xi}}(\vec{x})\right) + \breve{\vec{r}}(\vec{x}). \qedhere
	\end{align*}
\end{proof}

Similarly to PCA and KPCA, we have a best reconstruction map property for KPCA-RFF. This holds despite the fact that we relied on an approximate reproducing property.

\begin{theorem}\label{thm:rff-preimage-optimal}
	$\vec{\varphi}_{\text{dec}}^{\text{KPCA-RFF}}$ is the optimal affine reconstruction map in the least-squares sense.
\end{theorem}

This allows us to obtain the Tikhonov regularized pre-image map
\begin{equation}
	\vec{z} \mapsto \vec{\varphi}_{\text{dec}}^{\text{KPCA-RFF}}(\vec{z}; q, \delta) = \bar{\vec{x}} + \frac{1}{m} \tilde{X} \tilde{\Sigma}_X^\top V_q\left(\Lambda_q + \delta I\right)^{-1}\vec{z}.
\end{equation}

\begin{proof}
	Let $\vec{z} = \vec{\varphi}_{\text{enc}}^{\text{KPCA-RFF}}(\vec{x}; q) = V_q^\top \tilde{\vec{\xi}}(\vec{x})$. We show that $\vec{\varphi}_{\text{dec}}^{\text{KPCA-RFF}}$ minimizes the reconstruction error over the class of functions of the form $\vec{z} \mapsto \bar{\vec{x}} + W \vec{z}$. This optimization problem is equivalent to
	\begin{align*}
		\hat{W} &= \underset{W \in \R^{d\times q}}{\arg\min} \frac{1}{m} \sum_{i=1}^m \norm{\tilde{\vec{x}}^{(i)} - W \vec{z}^{(i)}}_2^2\\
		&= \underset{W \in \R^{d\times q}}{\arg\min} \frac{1}{m} \norm{\tilde{X} - W Z }_F^2,
	\end{align*}
	whose solution is given by $\hat{W} = \tilde{X} Z^\dagger$. Note that $Z = \begin{bmatrix}
		\vec{z}^{(1)} & \dots & \vec{z}^{(m)}
	\end{bmatrix} = V_q^\top \tilde{\Xi}_X$. Let $\tilde{\Xi}_X = R S Q^\top$ be the SVD of $\tilde{\Xi}_X$. Then one can show that $R = V$ and $\Lambda = \frac{1}{m} S^2$. It follows that $Z = S_q Q_q^\top$ and $Z^\dagger =  Q_q S_q^{-1}$. On the other hand,
	\begin{equation*}
		\frac{1}{m} \tilde{\Xi}_X^\top V_q \Lambda_q^{-1} = \frac{1}{m} Z^\top \Lambda_q^{-1} = \frac{1}{m} Q_q S_q \left(\frac{1}{m} S_q^2\right)^{-1} = Q_q S_q^{-1}.
	\end{equation*}
	Therefore, $\hat{W} = 	\frac{1}{m} \tilde{X} \tilde{\Xi}_X^\top V_q \Lambda_q^{-1}$.
\end{proof}

\section{Numerical experiments}\label{sec:experiments}

We evaluate the performance of the derived pre-image map on image denoising and dimensionality reduction tasks using both synthetic and real-world datasets. We use the USPS digits dataset~\cite{hull2002database} to compare the performance and computational efficiency of the proposed approach against existing methods. We additionally apply KPCA-RFF with our reconstruction map for dimensionality reduction to the MNIST~\cite{lecun2010mnist} and CIFAR-10~\cite{krizhevsky2009learning} datasets for which the cost of running exact KPCA would be prohibitive.  Performance is assessed using three metrics: the mean squared error (MSE), the peak signal-to-noise ratio (PSNR) in decibels (dB), and the structural similarity index measure (SSIM)~\cite{wang2004image}. We consider the following kernels: 
\begin{itemize}
	\item \emph{RBF kernel}: $k_{\gamma}(\vec{x}, \vec{y}) = \exp{\left(-\gamma \norm{\vec{x} - \vec{y}}_2^2 \right)}$, with $\gamma > 0$. The kernel bandwidth $\sigma_{\text{RBF}}$ is such that $\gamma = 0.5 /  \sigma_{\text{RBF}}^2 $.
	\item \emph{Polynomial kernel}:  $k_{p,\gamma,c}(\vec{x}, \vec{y}) = \left(\gamma \innerprod{\vec{x}}{\vec{y}} + c\right)^p$, with $\gamma > 0$, $c \geq 0$, and the degree $p$ is a positive integer. The \emph{linear kernel} corresponds to the case $p=\gamma=1$ and $c=0$.
	\item \emph{Sigmoid kernel}: $k_{\gamma,c}(\vec{x}, \vec{y}) = \tanh\left(\gamma \innerprod{\vec{x}}{\vec{y}} + c\right)$, with $\gamma > 0$ and $c \geq 0$. 
	\item \emph{Cosine kernel}: $k(\vec{x}, \vec{y}) = \frac{\innerprod{\vec{x}}{\vec{y}}}{\norm{\vec{x}}_2 \norm{\vec{y}}_2 }$.
\end{itemize}
All experiments are run in Python 3.12 on a standard laptop equipped with an Intel Core i7 processor (12th Gen) and 32\,GB of RAM.

\subsection{A toy experiment}

\begin{figure}
	\centering
	\includegraphics[width=0.6\linewidth]{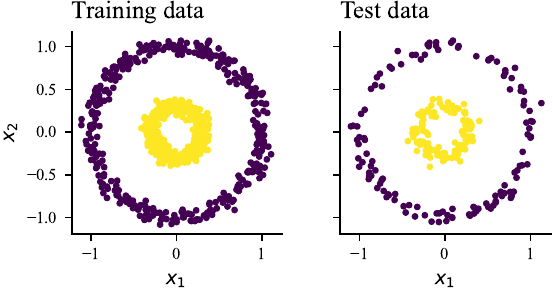}
	\caption{Two circles experiment. Training and test data.}
	\label{fig:two_cirlces_dataset}
\end{figure}

We first consider a synthetic dataset consisting of points on two nested circles of radii $1.0$ and $0.3$. The training data comprises $375$ evenly spaced points on each circle, perturbed by additive Gaussian noise with standard deviation $0.05$; test data is generated identically with $125$ points per circle (see Figure~\ref{fig:two_cirlces_dataset} for a visualization). This example is adapted from the \emph{scikit-learn} documentation.\footnote{\url{https://scikit-learn.org/stable/auto_examples/decomposition/plot_kernel_pca.html}} We use the RBF kernel with $\gamma = 10$ and $n = 200$ random features for KPCA-RFF.

\begin{figure}
	\centering
	\includegraphics[width=1.\linewidth]{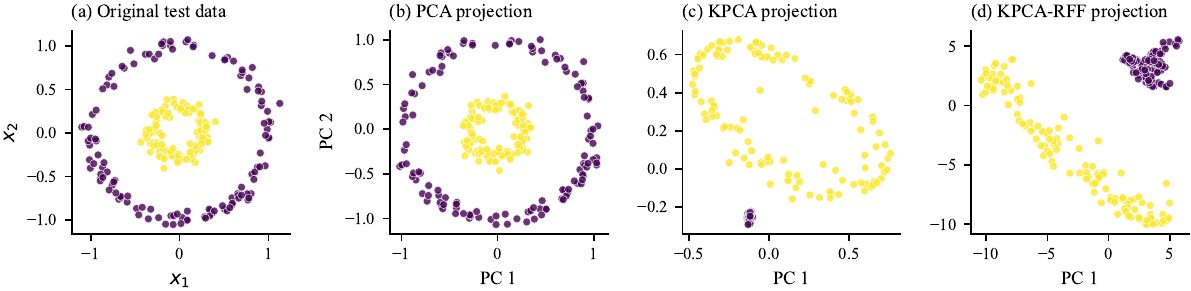}
	\caption{Two circles experiment. Projection onto the first two principal components using PCA, KPCA, and KPCA-RFF. In contrast to KPCA and KPCA-RFF, PCA fails to linearly separate data from the two circles.}
	\label{fig:two_cirlces_projection}
\end{figure}

\begin{figure}
	\centering
	\includegraphics[width=1.\linewidth]{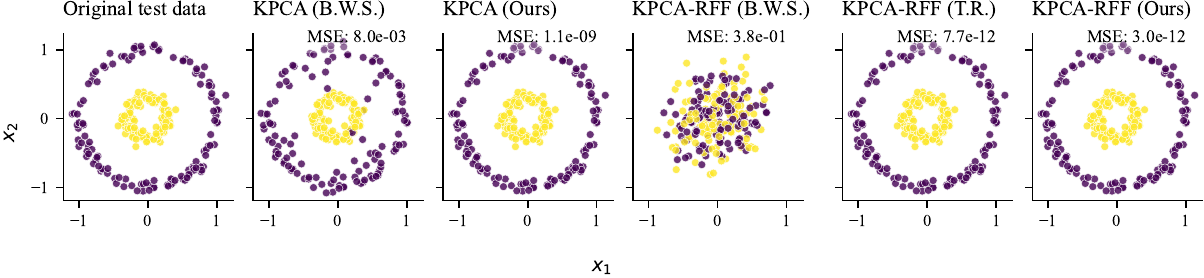}
	\caption{Two circles experiment. Reconstruction of the test data based on all the principal components for both KPCA and KPCA-RFF. Some baselines have been omitted due to space constraints. B.W.S. and T.R. correspond to algorithms from \cite{bakir2004learning} and \cite{tiddeman2023pre}, respectively.}
	\label{fig:two_cirlces_reconstruction}
\end{figure}

Figure~\ref{fig:two_cirlces_projection} shows the projection of the test data onto the first two principal components, illustrating that PCA fails to linearly separate the two circles, whereas both KPCA and KPCA-RFF succeed. As Figure~\ref{fig:two_cirlces_reconstruction} shows, the B.W.S. method~\cite{bakir2004learning} performs poorly on the reconstruction task for KPCA, exhibiting label switching, with points from one circle being mapped onto the other, and catastrophically fails for the RFF approximation.\!\footnote{We use \emph{scikit-learn}'s implementation of the B.W.S. method~\cite{bakir2004learning} for exact KPCA.} These results could not be improved by tuning the regularization parameter alone. On the other hand, the proposed algorithm and the T.R.~\cite{tiddeman2023pre} baseline (see Section~\ref{sec:kpca_rff_experiments} for more details) correctly preserve the geometry of the data. Due to space constraints, results for the other baseline methods are not included for this example.

\subsection{USPS digits}
\subsubsection{Setup}
Following~\cite{bakir2004learning, mika1998kernel}, we evaluate the proposed reconstruction map for dimensionality reduction and image denoising using the USPS dataset. We randomly select $2\ts000$ training, $1\ts000$ test, and $100$ validation examples using stratified sampling by digit label. We then add Gaussian noise of scale $\sigma$ to each training and test data point in the denoising case where we consider three noise levels: $\sigma \in \{0.1, 0.25, 0.5\}$. The validation set is used for tuning the ridge regularization parameter $\delta$ for the proposed method and the kernel ridge regression of~\cite{bakir2004learning}. In each scenario, $100$ log-spaced values in $[10^{-12}, 10^{0}]$ were considered for $\delta$ and the value with the lowest validation MSE was chosen. In the dimensionality reduction case (clean data), we set $\delta$ to zero for our method. We report results for $q=50$ and $q=100$ retained PCs and experiment with five different kernels: Gaussian RBF, polynomial, sigmoid, cosine, and linear. The kernel hyperparameters are set to $\gamma = 3.125\times 10^{-4}$ (such that the kernel bandwidth is $\sigma_{\text{RBF}} = 40$), $p=3$, and $c=1.0$.

\subsubsection{Performance comparison}

\begin{figure*}
	\centering
	\subfloat[Ground truth
	\label{fig:usps-gt}]{%
		\includegraphics[width=0.12\textwidth]
		{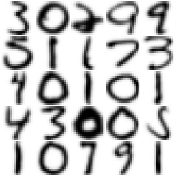}}%
	\hfill
	\subfloat[Noisy images
	\label{fig:usps-noisy}]{%
		\includegraphics[width=0.12\textwidth]
		{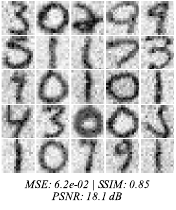}}%
	\hfill
	\subfloat[Ours
	\label{fig:usps-ours}]{%
		\includegraphics[width=0.12\textwidth]
		{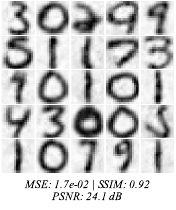}}%
	\hfill
	\subfloat[B.W.S.~\cite{bakir2004learning}
	\label{fig:usps-bakir}]{%
		\includegraphics[width=0.12\textwidth]
		{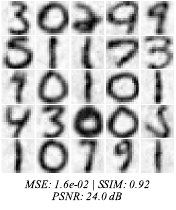}}%
	\hfill
	\subfloat[M. et al.~\cite{mika1998kernel}
	\label{fig:usps-mika}]{%
		\includegraphics[width=0.12\textwidth]
		{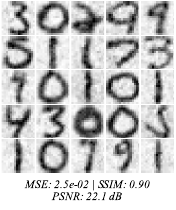}}%
	\hfill
	\subfloat[A.H.~\cite{abrahamsen2009input}
	\label{fig:usps-abrahamsen}]{%
	\includegraphics[width=0.12\textwidth]
	{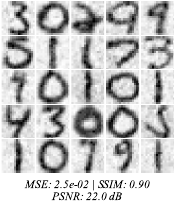}}%
	\hfill
	\subfloat[K.T.~\cite{kwok2004pre}
	\label{fig:usps-kwok}]{%
		\includegraphics[width=0.12\textwidth]
		{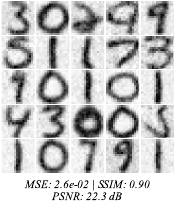}}%
	\hfill
	\subfloat[H.R.~\cite{honeine2011closed}
	\label{fig:usps-honeine}]{%
		\includegraphics[width=0.12\textwidth]
		{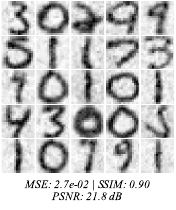}}%
	\caption{Image denoising on USPS digits. Representative results showing $25$ randomly selected examples from the test set obtained using the RBF kernel. The noise level is $\sigma = 0.25$. The panels compare the ground truth, noisy inputs, our method, and existing approaches. Metrics below each image grid report MSE, SSIM, and PSNR averaged over the full test set of $1000$ examples.}
	\label{fig:usps_denoising}
\end{figure*}

Figure~\ref{fig:usps_denoising} shows representative denoising outputs (for $\sigma = 0.25$) and associated metrics (measured on the entire test dataset) comparing our method against all the major preexisting approaches for RBF kernel and $q=100$. Our method performs comparably with the kernel ridge regression of~\cite{bakir2004learning} with the two methods achieving better quality denoising than the other approaches. For a broader assessment, Table~\ref{tab:reconstruction} reports comparison metrics on test data against the methods of~\cite{bakir2004learning, honeine2011closed} across all kernel choices. The other algorithms are not included since their implementations are kernel-specific and have historically been used for the RBF kernel alone.

We observe that, for the RBF and polynomial kernels, the approach of~\cite{bakir2004learning} is the best performing on clean data ($\sigma = 0$), achieving up to $43.6\%$ lower MSE compared to both our method and that of~\cite{honeine2011closed}, which performs comparably. This advantage is maintained in the low-noise regime ($\sigma = 0.1$) when the number of retained components is relatively small ($q = 50$)  with the method achieving a $16.2\%$ lower MSE compared to the second best. However, this performance gain over our method vanishes as the noise level is increased ($\sigma \in \{0.25, 0.5\}$) and when the number of retained components is large enough ($q=100$). In such settings, the H.R. approach~\cite{honeine2011closed} performs the worst. On the other hand, for the sigmoid kernel, the proposed algorithm matches the performance of the B.W.S. method across all settings, with both approaches performing better than the H.R. method in the denoising case.

In contrast, for the cosine and linear kernels, the B.W.S. approach degrades substantially across all noise levels and numbers of retained PCs, whereas our algorithm maintains consistent performance. The lowest achieved PSNR difference between the two methods is of about $10$~dB, with our method achieving up to $99.8\%$ lower MSE (linear kernel, $q= 100$). On the other hand, the conformal map H.R. approach completely fails for the cosine kernel, with the SSIM being zero on clean data. However, for the linear kernel, it obtains results that are comparable to ours in the clean and low-noise settings, but not when the noise level is high ($\sigma = 0.5$).

\input{usps_table_reconstruction.tex}

\subsubsection{Computational efficiency}

\begin{figure*}
	\centering
	\subfloat[Effect of test data size
	\label{fig:usps_efficiency_testsize}]{%
		\includegraphics[width=0.475\textwidth]
		{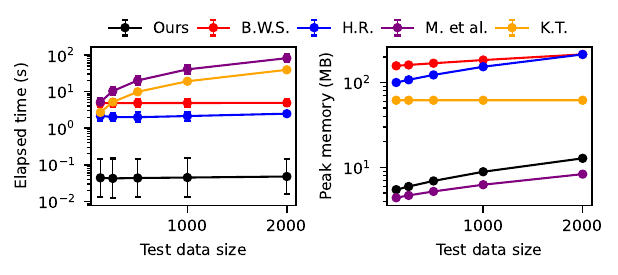}}%
	\hfill
	\subfloat[Effect of training data size
	\label{fig:usps_efficiency_trainsize}]{%
		\includegraphics[width=0.475\textwidth]
		{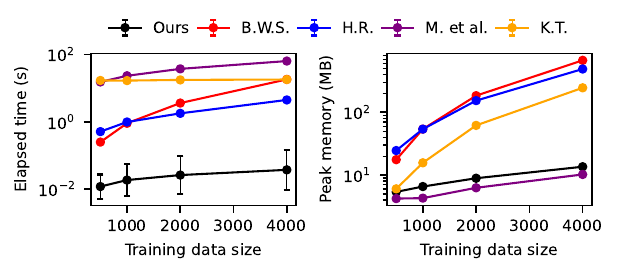}}%

	\caption{Computational requirements (elapsed time in seconds and peak memory usage in megabyte) for each KPCA pre-image algorithm vs.\ the number of (a) test and (b) training examples. We did not include the A.H. method~\cite{abrahamsen2009input} for the sake of readability. However, its complexity is comparable to M. et al.~\cite{mika1998kernel}.}
	\label{fig:usps_efficiency}
\end{figure*}

We further perform a systematic runtime and memory usage comparison of all the algorithms to evaluate their computational requirements. In particular, we investigate how the elapsed time (in seconds) and the peak memory usage (in MB) vary in terms of training and test data sizes. We use the RBF kernel and set the number of principal components to $50$. To ensure that our conclusions are robust, we run $10$ repetitions of each experiment and report the mean and standard deviation. Figure~\ref{fig:usps_efficiency_testsize} shows comparisons as the number of test examples is increased. Here, the training data size is fixed to $2\ts000$ examples. Across all settings, our algorithm is the most efficient. In particular, for $2\ts000$ test examples, its runtime is on average $160$x and $94$x lower compared to the B.W.S.~\cite{bakir2004learning} and H.R.~\cite{honeine2011closed} methods, respectively, with a $16.8$x lower peak memory usage. Only the  M. et al.~\cite{mika1998kernel} approach, which solves a separate optimization problem for each test example, has a marginally lower ($1.5$x) memory usage than ours. Furthermore, as Figure~\ref{fig:usps_efficiency_trainsize} shows, the achieved gains in efficiency are even more pronounced when the training data size $m$ is increased. Here, the number of test examples is fixed to $1\ts000$. Compared to the B.W.S.\ (and H.R.) method, our algorithm's runtime goes from being $29$x to $845$x (and $55$x to $210$x) lower as $m$ is increased from $500$ to $4\ts000$, while, at the same time, the peak memory usage goes from being $3.3$x to $49.7$x (and $4.6$x to $36.2$x) lower.

\subsection{Pre-image methods for KPCA-RFF}\label{sec:kpca_rff_experiments}

We now turn our attention to KPCA-RFF pre-image algorithms. As noted by the authors of \cite{tiddeman2023pre}, the pre-image problem for KPCA-RFF has not been widely investigated in the literature. In light of this, they proposed two approaches for input reconstruction: the first, which is inspired by the work of \cite{bakir2004learning}, learns a pre-image map by solving a linear ridge regression problem, whereas the second modifies the RFF approximation to make it invertible using a feature augmentation strategy. Theorem~\ref{thm:rff-preimage-optimal} shows that our KPCA-RFF decoder is structurally similar to this first approach. However, our method has the benefit of being derived from first principles, which also eliminates the need for a second pass through the data when learning the pre-image map. Using the RBF kernel, we compare our algorithm against this method (henceforth denoted by T.R.) and the RFF approximation of the B.W.S. algorithm~\cite{bakir2004learning}.

\subsubsection{USPS dataset}
We start by systematically comparing our algorithm with these two baseline methods using the USPS digits dataset. In particular, we investigate the effect of the kernel bandwidth, number of random features, training data size, and the number of retained principal components on the performance and computational efficiency of each algorithm. We consider an image denoising setting, with noise scale $\sigma = 0.25$.

\begin{figure*}
	\centering
	\subfloat[
	\label{fig:usps_bandwidth_rff}]{%
		\includegraphics[width=0.485\textwidth]
		{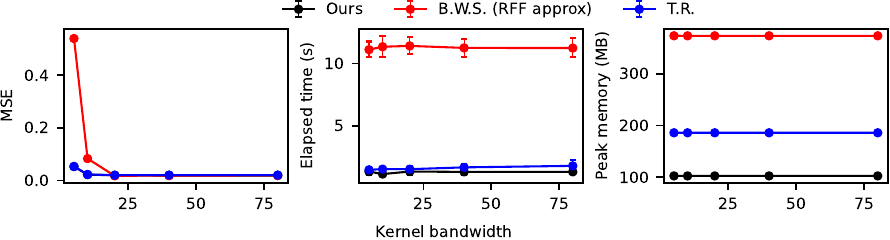}}%
	\hfill
	\subfloat[
	\label{fig:usps_num_rffs_rff}]{%
		\includegraphics[width=0.485\textwidth]
		{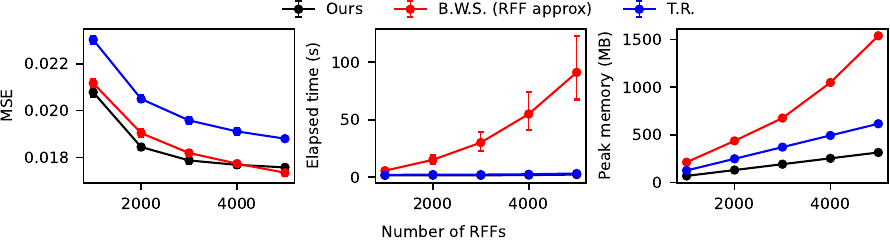}}%
	\\[-1ex]
	\subfloat[
	\label{fig:usps_trainsize_rff}]{%
	\includegraphics[width=0.485\textwidth]
	{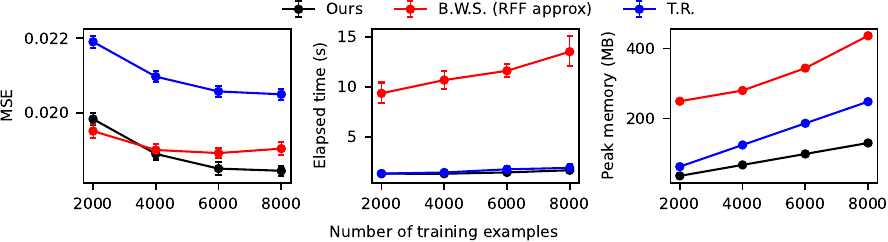}}%
	\hfill
	\subfloat[
	\label{fig:usps_num_comp_rff}]{%
		\includegraphics[width=0.485\textwidth]
		{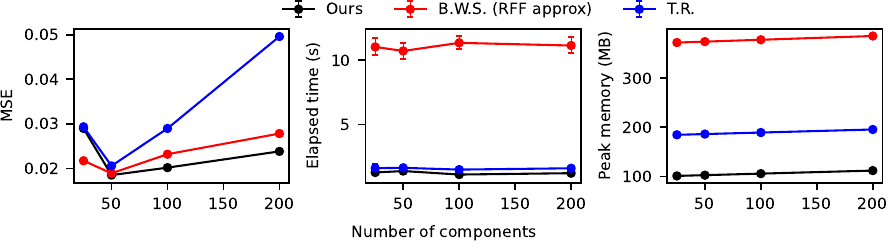}}%

	\caption{Comparison of KPCA-RFF pre-image algorithms using the USPS dataset. We only use the test data MSE as a performance metric. However, the results are similar for the SSIM and PNSR. We further report the computational requirements (elapsed time in seconds and peak memory usage in megabyte) of each algorithm, where T.R. refers to the first method of \cite{tiddeman2023pre}. (a) Effect of kernel bandwidth for fixed $n_{\text{train}} = 6\ts000, n_{\text{test}} = 2\ts000$, $n_{\text{RFFs}} = 2\ts000$, and number of components $q = 50$. (b) Effect of the number of RFFs for fixed $\sigma_{\text{RBF}} = 50, n_{\text{train}} = 8\ts000, n_{\text{test}} = 1\ts000, q=50$. (c) Effect of training data size for fixed $\sigma_{\text{RBF}} = 50, n_{\text{RFFs}} = 2\ts000, n_{\text{test}} = 1\ts000, q = 50$. (d) Effect of the number of principal components for fixed $\sigma_{\text{RBF}} = 50, n_{\text{train}} = 6\ts000, n_{\text{test}} = 2\ts000$, and $n_{\text{RFFs}} = 2\ts000$.}
	\label{fig:usps_kpca_rff}
\end{figure*}

Figure~\ref{fig:usps_kpca_rff} shows that our algorithm not only performs comparably or better than the B.W.S. method overall, but is also more efficient and has lower memory usage. Notably, as Figure~\ref{fig:usps_bandwidth_rff} indicates, the B.W.S. approach performs significantly worse when the kernel bandwidth is not large enough. Additionally, our method achieves lower MSEs and memory usage than the T.R. algorithm, whilst the two methods have comparable runtime. This worse performance of the T.R. approach, which is the most pronounced when the number of PCs is high (see Figure~\ref{fig:usps_num_comp_rff}), might be explained by the fact that the constant term is also learned from data, whereas our derivation fixes it to the mean of the training data.

\subsubsection{Large scale experiments on MNIST and CIFAR-10}
\input{mnist_table_reconstruction.tex}
\input{cifar10_table_reconstruction.tex}
To evaluate the performance on larger datasets, we apply KPCA-RFF to MNIST ($28\times 28$ grayscale handwritten digits) and CIFAR-10 ($32\times 32$ RGB images across $10$ classes) for dimensionality reduction and compare the performance and computational costs of the three pre-image methods. We use the canonical train--test splits: $60\ts000 / 10\ts000$ for MNIST and $50\ts000/10\ts000$ for CIFAR-10. It is worth noting that these datasets were also considered in the original KPCA-RFF paper~\cite{lopez2014randomized}. However, no details were provided regarding the reconstruction map beyond noting that it is nonlinear. For these experiments, we set the number of random features to $n = 2\ts000$ and report averaged results over $10$ runs.

Tables~\ref{tab:mnist_results}--\ref{tab:cifar10_results} show performance and compute metrics for MNIST and CIFAR-10 for different values of $\sigma_{\text{RBF}}$ with $q=100$ PCs, respectively. We observe that the B.W.S. method performs poorly for small kernel bandwidths, but it can outperform the other methods when $\sigma_{\text{RBF}}$ is large enough. This is similar to the USPS results in Figure~\ref{fig:usps_bandwidth_rff}. Additionally, it is the slowest method overall. On the other hand, our approach is the most computationally efficient and maintains the same performance as the T.R. algorithm across all settings.

\begin{figure}
	\centering
	\includegraphics[width=1\linewidth]{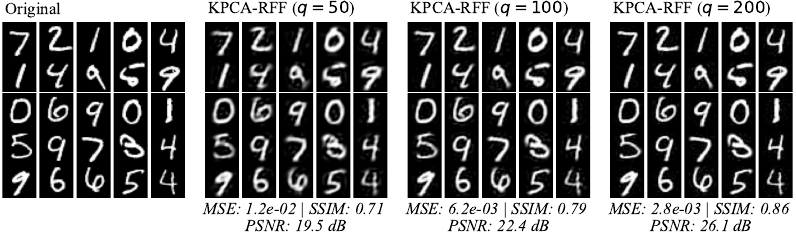}
	\caption{KPCA-RFF autoencoder on MNIST. Reconstruction of unseen images using the proposed method. Metrics are averaged over the full test set.}
	\label{fig:mnist_reconstruction}
\end{figure}

\begin{figure}
	\centering
	\includegraphics[width=1\linewidth]{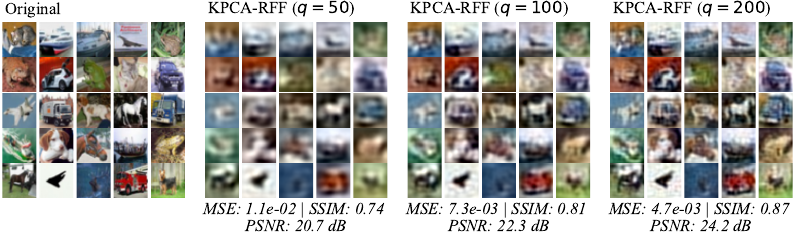}
	\caption{KPCA-RFF autoencoder on CIFAR-10. Reconstruction of unseen images using the proposed method. Metrics are averaged over the full test set.}
	\label{fig:cifar10_reconstruction}
\end{figure}

Figure \ref{fig:mnist_reconstruction} shows representative reconstruction results for the MNIST data for varying latent-space dimension -- with metrics averaged over the full test set -- obtained using the proposed algorithm. Despite the $784$--dimensional input space, a latent dimension of $50$ already yields high-quality reconstructions. Similar behavior is observed for the $3\ts072$--dimensional CIFAR-10 data, see Figure \ref{fig:cifar10_reconstruction}.

\section{Discussion}\label{sec:discussion}

In this work, we proposed a solution to the longstanding pre-image problem in KPCA that follows naturally from the reproducing property when the associated RKHS contains linear functions or the kernel is universal -- an assumption that holds for standard kernels used in machine learning. The obtained reconstruction map is optimal in the least-squares sense and consistent with standard PCA's decoder. Furthermore, by introducing the concept of the \emph{RFF approximate reproducing property}, we showed that the proposed method naturally extends to the random Fourier features approximation of KPCA. Numerical results demonstrate that, across kernel choices and applications, the proposed method performs comparably or significantly better than the state-of-the-art baselines while being more computationally efficient.

An important limitation of our approach is the restriction to vectorial data. Extending our results to other data types such as strings and graphs is non-trivial and worth further investigation. Other avenues for future research include extending this work to the related \emph{distributional pre-image problem}, where the goal is to recover a probability distribution given its kernel mean embedding, see, e.g.~\cite{kanagawa2014recovering, muandet2017kernel, schuster2020kernel}. Another direction is to adapt the proposed method to the Nystr\"om approximation of KPCA~\cite{zhang2008improved, sterge2020gain, sterge2022statistical}. Finally, a rigorous convergence analysis for the KPCA-RFF extension is needed.

\section*{Code and data availability}
The Python code to reproduce the results presented in this paper is available at \href{https://github.com/J-PNtwali/edmd-kdl}{github.com/J-PNtwali/kpca\_preimage}. All the datasets used are publicly accessible.

\section*{Acknowledgments}
	J.-P.\ was supported by the EPSRC Centre for Doctoral Training in Mathematical Modelling, Analysis and Computation (MAC-MIGS) funded by the UK Engineering and Physical Sciences Research Council (grant EP/S023291/1), Heriot--Watt University and The University of Edinburgh.


\bibliographystyle{IEEEtran}
\bibliography{kPCA_PreImage}


\newpage

%
%
%
%

\vfill

\end{document}

%% file: usps_table_reconstruction.tex
\begin{table*}
\centering
\caption{USPS digits data: reconstruction metrics across kernels, numbers of retained components, noise levels, and methods (B.W.S. corresponds to the method of \cite{bakir2004learning} and H.R. to that of \cite{honeine2011closed}). Bold values indicate the best result for each metric and noise level, while underlined values indicate ties at the displayed precision.}
\label{tab:reconstruction}
\small
\setlength{\tabcolsep}{4.pt}
\renewcommand{\arraystretch}{1.08}
\begin{tabular}{ll@{\hspace{0.8em}}ccc@{\hspace{0.8em}}ccc@{\hspace{0.8em}}ccc@{\hspace{0.8em}}ccc}
\toprule
\multicolumn{2}{c}{} & \multicolumn{3}{c}{$\sigma=0$} & \multicolumn{3}{c}{$\sigma=0.1$} & \multicolumn{3}{c}{$\sigma=0.25$} & \multicolumn{3}{c}{$\sigma=0.5$} \\
\cmidrule(lr){3-5} \cmidrule(lr){6-8} \cmidrule(lr){9-11} \cmidrule(lr){12-14}
Kernel & Metric & B.W.S. & H.R. & Ours & B.W.S. & H.R. & Ours & B.W.S. & H.R. & Ours & B.W.S. & H.R. & Ours \\
\midrule
\multicolumn{14}{l}{\textit{$q = 50$}} \\
\addlinespace[2.5pt]
\multirow{3}{*}{RBF} & MSE ($\downarrow$) & \textbf{3.1e-03} & 5.5e-03 & 5.5e-03 & \textbf{6.2e-03} & 7.5e-03 & 7.4e-03 & \underline{1.7e-02} & 1.9e-02 & \underline{1.7e-02} & \underline{4.1e-02} & 6.1e-02 & \underline{4.1e-02} \\
 & SSIM ($\uparrow$) & \textbf{0.99} & 0.98 & 0.98 & \underline{0.97} & \underline{0.97} & \underline{0.97} & \underline{0.92} & \underline{0.92} & \underline{0.92} & \underline{0.82} & 0.80 & \underline{0.82} \\
 & PSNR ($\uparrow$) & \textbf{33.0} & 29.8 & 29.8 & \textbf{28.6} & 27.8 & 27.9 & \underline{24.1} & 23.4 & \underline{24.1} & \underline{20.1} & 18.3 & \underline{20.1} \\
\addlinespace[2.5pt]
\multirow{3}{*}{Poly} & MSE ($\downarrow$) & \textbf{3.1e-03} & 5.5e-03 & 5.5e-03 & \textbf{6.2e-03} & 7.5e-03 & 7.4e-03 & \underline{1.7e-02} & 1.9e-02 & \underline{1.7e-02} & \underline{4.2e-02} & 6.1e-02 & \underline{4.2e-02} \\
 & SSIM ($\uparrow$) & \textbf{0.99} & 0.98 & 0.98 & \underline{0.97} & \underline{0.97} & \underline{0.97} & \underline{0.92} & \underline{0.92} & \underline{0.92} & \underline{0.82} & 0.80 & \underline{0.82} \\
 & PSNR ($\uparrow$) & \textbf{33.0} & 29.9 & 29.9 & \textbf{28.6} & 27.8 & 27.9 & \underline{24.1} & 23.4 & \underline{24.1} & \underline{20.1} & 18.2 & \underline{20.1} \\
\addlinespace[2.5pt]
\multirow{3}{*}{Sigmoid} & MSE ($\downarrow$) & \underline{5.3e-03} & \underline{5.3e-03} & \underline{5.3e-03} & \underline{7.4e-03} & 7.5e-03 & \underline{7.4e-03} & \underline{1.7e-02} & 1.9e-02 & \underline{1.7e-02} & \underline{4.2e-02} & 6.2e-02 & \underline{4.2e-02} \\
 & SSIM ($\uparrow$) & \underline{0.99} & \underline{0.99} & \underline{0.99} & \underline{0.97} & \underline{0.97} & \underline{0.97} & \underline{0.92} & \underline{0.92} & \underline{0.92} & \underline{0.82} & 0.79 & \underline{0.82} \\
 & PSNR ($\uparrow$) & \underline{29.9} & \underline{29.9} & \underline{29.9} & \underline{27.9} & 27.8 & \underline{27.9} & \underline{24.0} & 23.3 & \underline{24.0} & \underline{20.1} & 18.2 & \underline{20.1} \\
\addlinespace[2.5pt]
\multirow{3}{*}{Cosine} & MSE ($\downarrow$) & 3.0e-01 & 1.3e+05 & \textbf{6.2e-03} & 3.1e-01 & 2.1e+02 & \textbf{8.1e-03} & 3.2e-01 & 7.9e+01 & \textbf{1.7e-02} & 3.7e-01 & 4.5e+00 & \textbf{4.1e-02} \\
 & SSIM ($\uparrow$) & 0.25 & 0.00 & \textbf{0.97} & 0.24 & 0.08 & \textbf{0.96} & 0.23 & 0.12 & \textbf{0.92} & 0.19 & 0.31 & \textbf{0.82} \\
 & PSNR ($\uparrow$) & 11.7 & -37.9 & \textbf{29.1} & 11.6 & -10.2 & \textbf{27.5} & 11.4 & -5.9 & \textbf{24.0} & 10.6 & 4.8 & \textbf{20.1} \\
\addlinespace[2.5pt]
\multirow{3}{*}{Linear} & MSE ($\downarrow$) & 3.5e-01 & \underline{5.3e-03} & \underline{5.3e-03} & 3.5e-01 & 7.4e-03 & \textbf{7.3e-03} & 3.7e-01 & 1.9e-02 & \textbf{1.7e-02} & 4.1e-01 & 6.2e-02 & \textbf{4.2e-02} \\
 & SSIM ($\uparrow$) & 0.14 & \underline{0.99} & \underline{0.99} & 0.13 & \underline{0.97} & \underline{0.97} & 0.13 & \underline{0.92} & \underline{0.92} & 0.12 & 0.80 & \textbf{0.82} \\
 & PSNR ($\uparrow$) & 10.6 & \underline{30.0} & \underline{30.0} & 10.5 & 27.8 & \textbf{27.9} & 10.4 & 23.4 & \textbf{24.0} & 9.9 & 18.2 & \textbf{20.1} \\
\midrule
\multicolumn{14}{l}{\textit{$q = 100$}} \\
\addlinespace[2.5pt]
\multirow{3}{*}{RBF} & MSE ($\downarrow$) & \textbf{6.6e-04} & 1.1e-03 & 1.1e-03 & \underline{4.1e-03} & 4.7e-03 & \underline{4.1e-03} & \underline{1.6e-02} & 2.7e-02 & \underline{1.6e-02} & \underline{4.5e-02} & 1.0e-01 & \underline{4.5e-02} \\
 & SSIM ($\uparrow$) & \underline{1.00} & \underline{1.00} & \underline{1.00} & \underline{0.98} & \underline{0.98} & \underline{0.98} & \underline{0.92} & 0.90 & \underline{0.92} & \underline{0.81} & 0.74 & \underline{0.81} \\
 & PSNR ($\uparrow$) & \textbf{39.8} & 37.0 & 37.0 & \textbf{30.1} & 29.4 & 30.0 & \underline{24.0} & 21.8 & \underline{24.0} & \underline{19.7} & 15.9 & \underline{19.7} \\
\addlinespace[2.5pt]
\multirow{3}{*}{Poly} & MSE ($\downarrow$) & \textbf{6.5e-04} & 1.1e-03 & 1.1e-03 & \underline{4.1e-03} & 4.7e-03 & \underline{4.1e-03} & \underline{1.7e-02} & 2.7e-02 & \underline{1.7e-02} & \underline{4.5e-02} & 1.0e-01 & \underline{4.5e-02} \\
 & SSIM ($\uparrow$) & \underline{1.00} & \underline{1.00} & \underline{1.00} & \underline{0.98} & \underline{0.98} & \underline{0.98} & \underline{0.92} & 0.90 & \underline{0.92} & \underline{0.81} & 0.74 & \underline{0.81} \\
 & PSNR ($\uparrow$) & \textbf{39.8} & 37.0 & 37.0 & \underline{30.0} & 29.4 & \underline{30.0} & \underline{24.0} & 21.7 & \underline{24.0} & \underline{19.7} & 15.9 & \underline{19.7} \\
\addlinespace[2.5pt]
\multirow{3}{*}{Sigmoid} & MSE ($\downarrow$) & \underline{6.7e-04} & \underline{6.7e-04} & \underline{6.7e-04} & \underline{4.2e-03} & 5.1e-03 & \underline{4.2e-03} & \underline{1.7e-02} & 2.8e-02 & \underline{1.7e-02} & \underline{4.5e-02} & 1.1e-01 & \underline{4.5e-02} \\
 & SSIM ($\uparrow$) & \underline{1.00} & \underline{1.00} & \underline{1.00} & \underline{0.98} & \underline{0.98} & \underline{0.98} & \underline{0.92} & 0.90 & \underline{0.92} & \underline{0.80} & 0.74 & \underline{0.80} \\
 & PSNR ($\uparrow$) & \underline{38.9} & \underline{38.9} & \underline{38.9} & \underline{29.9} & 29.0 & \underline{29.9} & \underline{23.9} & 21.6 & \underline{23.9} & \underline{19.7} & 15.8 & \underline{19.7} \\
\addlinespace[2.5pt]
\multirow{3}{*}{Cosine} & MSE ($\downarrow$) & 3.0e-01 & 8.6e+05 & \textbf{1.3e-03} & 3.0e-01 & 1.0e+03 & \textbf{4.8e-03} & 3.2e-01 & 1.6e+02 & \textbf{1.7e-02} & 4.1e-01 & 6.8e+00 & \textbf{4.5e-02} \\
 & SSIM ($\uparrow$) & 0.25 & 0.00 & \textbf{0.99} & 0.25 & 0.03 & \textbf{0.97} & 0.22 & 0.08 & \textbf{0.92} & 0.17 & 0.25 & \textbf{0.80} \\
 & PSNR ($\uparrow$) & 11.8 & -45.3 & \textbf{36.1} & 11.7 & -17.0 & \textbf{29.5} & 11.2 & -9.4 & \textbf{24.0} & 10.0 & 2.8 & \textbf{19.8} \\
\addlinespace[2.5pt]
\multirow{3}{*}{Linear} & MSE ($\downarrow$) & 3.5e-01 & \underline{5.9e-04} & \underline{5.9e-04} & 4.0e-01 & 4.9e-03 & \textbf{4.1e-03} & 3.7e-01 & 2.8e-02 & \textbf{1.7e-02} & 4.6e-01 & 1.1e-01 & \textbf{4.5e-02} \\
 & SSIM ($\uparrow$) & 0.14 & \underline{1.00} & \underline{1.00} & 0.17 & \underline{0.98} & \underline{0.98} & 0.13 & 0.90 & \textbf{0.92} & 0.10 & 0.74 & \textbf{0.80} \\
 & PSNR ($\uparrow$) & 10.6 & \underline{39.5} & \underline{39.5} & 10.8 & 29.2 & \textbf{30.0} & 10.3 & 21.7 & \textbf{24.0} & 9.5 & 15.8 & \textbf{19.7} \\
\bottomrule
\end{tabular}
\end{table*}

%% file: mnist_table_reconstruction.tex
\begin{table}[ht]
\centering
\caption{MNIST: Reconstruction quality, runtime and memory.}
\label{tab:mnist_results}
\begin{tabular}{clccc}
\toprule
$\sigma_{\text{RBF}}$ & Measure & B.W.S.~\cite{bakir2004learning} & Ours & T.R.~\cite{tiddeman2023pre} \\
\midrule
5 & MSE & $1.2\times 10^{-1}$ & $\mathbf{1.4\times 10^{-2}}$ & $\mathbf{1.4\times 10^{-2}}$ \\
 & SSIM & 0.10 & \textbf{0.67} & \textbf{0.67} \\
 & PSNR (dB) & 9.8 & \textbf{19.1} & \textbf{19.1} \\
 & Time (s) & 30.3 & \textbf{4.0} & 10.1 \\
 & Memory (MB) & 1880.0 & \textbf{963.0} & 1878.5 \\
\midrule
10 & MSE & $3.2\times 10^{-2}$ & $\mathbf{7.6\times 10^{-3}}$ & $\mathbf{7.6\times 10^{-3}}$ \\
 & SSIM & 0.49 & \textbf{0.76} & \textbf{0.76} \\
 & PSNR (dB) & 15.9 & \textbf{21.5} & \textbf{21.5} \\
 & Time (s) & 27.0 & \textbf{3.2} & 9.3 \\
 & Memory (MB) & 1880.0 & \textbf{963.0} & 1878.5 \\
\midrule
20 & MSE & $\mathbf{5.2\times 10^{-3}}$ & $6.3\times 10^{-3}$ & $6.3\times 10^{-3}$ \\
 & SSIM & \textbf{0.82} & 0.79 & 0.79 \\
 & PSNR (dB) & \textbf{23.4} & 22.4 & 22.4 \\
 & Time (s) & 27.0 & \textbf{3.3} & 8.9 \\
 & Memory (MB) & 1880.0 & \textbf{963.0} & 1878.5 \\
\midrule
40 & MSE & $\mathbf{4.7\times 10^{-3}}$ & $6.2\times 10^{-3}$ & $6.2\times 10^{-3}$ \\
 & SSIM & \textbf{0.83} & 0.79 & 0.79 \\
 & PSNR (dB) & \textbf{23.8} & 22.4 & 22.4 \\
 & Time (s) & 35.4 & \textbf{6.9} & 11.3 \\
 & Memory (MB) & 1880.0 & \textbf{963.0} & 1878.5 \\
\bottomrule
\end{tabular}
\end{table}

%% file: cifar10_table_reconstruction.tex
\begin{table}[ht]
\centering
\caption{CIFAR-10: Reconstruction quality, runtime and memory.}
\label{tab:cifar10_results}
\begin{tabular}{clccc}
\toprule
$\sigma_{\text{RBF}}$ & Measure & B.W.S.~\cite{bakir2004learning} & Ours & T.R.~\cite{tiddeman2023pre} \\
\midrule
5 & MSE & $1.9\times 10^{-1}$ & $\mathbf{4.1\times 10^{-2}}$ & $\mathbf{4.1\times 10^{-2}}$ \\
 & SSIM & 0.06 & \textbf{0.37} & \textbf{0.37} \\
 & PSNR (dB) & 8.8 & \textbf{15.3} & \textbf{15.3} \\
 & Time (s) & 38.8 & \textbf{5.3} & 13.6 \\
 & Memory (MB) & 1605.0 & \textbf{1212.5} & 1565.7 \\
\midrule
10 & MSE & $8.6\times 10^{-2}$ & $\mathbf{1.6\times 10^{-2}}$ & $\mathbf{1.6\times 10^{-2}}$ \\
 & SSIM & 0.30 & \textbf{0.68} & \textbf{0.68} \\
 & PSNR (dB) & 12.2 & \textbf{19.0} & \textbf{19.0} \\
 & Time (s) & 34.8 & \textbf{4.8} & 12.4 \\
 & Memory (MB) & 1605.0 & \textbf{1212.5} & 1565.7 \\
\midrule
20 & MSE & $1.0\times 10^{-2}$ & $\mathbf{9.0\times 10^{-3}}$ & $\mathbf{9.0\times 10^{-3}}$ \\
 & SSIM & 0.77 & \textbf{0.78} & \textbf{0.78} \\
 & PSNR (dB) & 20.9 & \textbf{21.3} & \textbf{21.3} \\
 & Time (s) & 36.3 & \textbf{5.0} & 13.4 \\
 & Memory (MB) & 1605.0 & \textbf{1212.5} & 1565.7 \\
\midrule
40 & MSE & $\mathbf{7.3\times 10^{-3}}$ & $\mathbf{7.3\times 10^{-3}}$ & $\mathbf{7.3\times 10^{-3}}$ \\
 & SSIM & \textbf{0.81} & \textbf{0.81} & \textbf{0.81} \\
 & PSNR (dB) & \textbf{22.4} & 22.3 & 22.3 \\
 & Time (s) & 39.2 & \textbf{8.3} & 13.8 \\
 & Memory (MB) & 1605.0 & \textbf{1212.5} & 1565.7 \\
\bottomrule
\end{tabular}
\end{table}

%% file: kPCA_PreImage.bib
@inproceedings{lopez2014randomized,
  title={Randomized nonlinear component analysis},
  author={Lopez-Paz, David and Sra, Suvrit and Smola, Alex and Ghahramani, Zoubin and Sch{\"o}lkopf, Bernhard},
  booktitle={International {C}onference on {M}achine {L}earning},
  pages={1359--1367},
  year={2014},
  organization={PMLR}
}

@article{scholkopf1998nonlinear,
	title={Nonlinear component analysis as a kernel eigenvalue problem},
	author={Sch{\"o}lkopf, Bernhard and Smola, Alexander and M{\"u}ller, Klaus-Robert},
	journal={Neural computation},
	volume={10},
	number={5},
	pages={1299--1319},
	year={1998},
	publisher={MIT Press}
}

@book{scholkopf2002learning,
	title={Learning with {K}ernels: {S}upport {V}ector {M}achines, {R}egularization, {O}ptimization, and {B}eyond},
	author={Sch{\"o}lkopf, Bernhard and Smola, Alexander J},
	year={2002},
	publisher={MIT {P}ress}
}

@article{bakir2004learning,
	title={Learning to find pre-images},
	author={Bak{\i}r, G{\"o}khan H and Weston, Jason and Sch{\"o}lkopf, Bernhard},
	journal={{A}dvances in {N}eural {I}nformation {P}rocessing {S}ystems},
	volume={16},
	number={7},
	pages={449--456},
	year={2004}
}

@inproceedings{schuster2020kernel,
	title={Kernel conditional density operators},
	author={Schuster, Ingmar and Mollenhauer, Mattes and Klus, Stefan and Muandet, Krikamol},
	booktitle={International {C}onference on {A}rtificial {I}ntelligence and {S}tatistics},
	pages={993--1004},
	year={2020},
	organization={PMLR}
}

@inproceedings{tiddeman2023pre,
	title={Pre-image calculation for random {F}ourier feature kernel machines},
	author={Tiddeman, Bernard and Robinson, Will},
	booktitle={UK {W}orkshop on {C}omputational {I}ntelligence},
	pages={433--444},
	year={2023},
	organization={Springer}
}

@article{sterge2022statistical,
	title={Statistical optimality and computational efficiency of {N}ystr{\"o}m kernel {PCA}},
	author={Sterge, Nicholas and Sriperumbudur, Bharath {K}},
	journal={Journal of {M}achine {L}earning {R}esearch},
	volume={23},
	number={337},
	pages={1--32},
	year={2022}
}

@article{rahimi2007random,
	title={Random features for large-scale kernel machines},
	author={Rahimi, Ali and Recht, Benjamin},
	journal={{A}dvances in {N}eural {I}nformation {P}rocessing {S}ystems},
	volume={20},
	year={2007}
}

@article{mika1998kernel,
	title={Kernel {PCA} and de-noising in feature spaces},
	author={Mika, Sebastian and Sch{\"o}lkopf, Bernhard and Smola, Alex and M{\"u}ller, Klaus-Robert and Scholz, Matthias and R{\"a}tsch, Gunnar},
	journal={{A}dvances in {N}eural {I}nformation {P}rocessing {S}ystems},
	volume={11},
	year={1998}
}

@article{scikit-learn,
	title={Scikit-learn: Machine Learning in {P}ython},
	author={Pedregosa, F. and Varoquaux, G. and Gramfort, A. and Michel, V.
	and Thirion, B. and Grisel, O. and Blondel, M. and Prettenhofer, P.
	and Weiss, R. and Dubourg, V. and Vanderplas, J. and Passos, A. and
	Cournapeau, D. and Brucher, M. and Perrot, M. and Duchesnay, E.},
	journal={Journal of {M}achine {L}earning {R}esearch},
	volume={12},
	pages={2825--2830},
	year={2011}
}

@article{hull2002database,
	title={A database for handwritten text recognition research},
	author={Hull, Jonathan J.},
	journal={{IEEE} {T}ransactions on {P}attern {A}nalysis and {M}achine {I}ntelligence},
	volume={16},
	number={5},
	pages={550--554},
	year={2002},
	publisher={IEEE}
}

@article{lecun2010mnist,
	title={{MNIST} handwritten digit database},
	author={Le{C}un, Yann and Cortes, Corinna and Burges, CJ},
	journal={{ATT} {L}abs [Online]. Available: http://yann.lecun.com/exdb/mnist},
	volume={2},
	year={2010}
}

@techreport{krizhevsky2009learning,
	title       = {Learning Multiple Layers of Features from Tiny Images},
	author      = {Krizhevsky, Alex},
	institution = {University of Toronto},
	year        = {2009},
	url         = {https://www.cs.toronto.edu/~kriz/learning-features-2009-TR.pdf}
}

@article{wang2004image,
	title={Image quality assessment: from error visibility to structural similarity},
	author={Wang, Zhou and Bovik, Alan C and Sheikh, Hamid R and Simoncelli, Eero P},
	journal={{IEEE} {T}ransactions on {I}mage {P}rocessing},
	volume={13},
	number={4},
	pages={600--612},
	year={2004},
	publisher={IEEE}
}

@article{baldi1989neural,
	title={Neural networks and principal component analysis: {L}earning from examples without local minima},
	author={Baldi, Pierre and Hornik, Kurt},
	journal={Neural networks},
	volume={2},
	number={1},
	pages={53--58},
	year={1989},
	publisher={Elsevier}
}

@article{pearson1901pca,
	title={On lines and planes of closest fit to systems of points in space},
	author={Pearson, Karl},
	journal={The {L}ondon, {E}dinburgh, and {D}ublin {P}hilosophical {M}agazine and {J}ournal of {S}cience},
	volume={2},
	number={11},
	pages={559--572},
	year={1901},
	publisher={Taylor \& {F}rancis}
}

@article{hotelling1933analysis,
	title={Analysis of a complex of statistical variables into principal components.},
	author={Hotelling, Harold},
	journal={Journal of {E}ducational {P}sychology},
	volume={24},
	number={6},
	pages={417},
	year={1933},
	publisher={Warwick \& {Y}ork}
}

@article{tipping1999probabilistic,
	title={Probabilistic principal component analysis},
	author={Tipping, Michael E and Bishop, Christopher M},
	journal={Journal of the {R}oyal {S}tatistical {S}ociety {S}eries {B}: {S}tatistical {M}ethodology},
	volume={61},
	number={3},
	pages={611--622},
	year={1999},
	publisher={Oxford {U}niversity {P}ress}
}

@article{roweis1997algorithms,
	title={{EM} algorithms for {PCA} and {SPCA}},
	author={Roweis, Sam},
	journal={Advances in {N}eural {I}nformation {P}rocessing {S}ystems},
	volume={10},
	year={1997}
}

@article{lucas2019don,
	title={Don't {B}lame the {ELBO}! {A} {L}inear {VAE} {P}erspective on {P}osterior {C}ollapse},
	author={Lucas, James and Tucker, George and Grosse, Roger B and Norouzi, Mohammad},
	journal={Advances in {N}eural {I}nformation {P}rocessing {S}ystems},
	volume={32},
	year={2019}
}

@article{hinton2006reducing,
	title={Reducing the dimensionality of data with neural networks},
	author={Hinton, Geoffrey E and Salakhutdinov, Ruslan R},
	journal={Science},
	volume={313},
	number={5786},
	pages={504--507},
	year={2006},
	publisher={American {A}ssociation for the {A}dvancement of {S}cience}
}

@inproceedings{sterge2020gain,
	title={Gain with no pain: {E}fficiency of kernel-{PCA} by {N}ystr{\"o}m sampling},
	author={Sterge, Nicholas and Sriperumbudur, Bharath and Rosasco, Lorenzo and Rudi, Alessandro},
	booktitle={{I}nternational {C}onference on {A}rtificial {I}ntelligence and {S}tatistics},
	pages={3642--3652},
	year={2020},
	organization={{PMLR}}
}

@inproceedings{zhang2008improved,
	title={Improved {N}ystr{\"o}m low-rank approximation and error analysis},
	author={Zhang, Kai and Tsang, Ivor W and Kwok, James T},
	booktitle={{P}roceedings of the 25th {I}nternational {C}onference on {M}achine {L}earning},
	pages={1232--1239},
	year={2008}
}

@article{sriperumbudur2022approximate,
	title={Approximate kernel {PCA}: {C}omputational versus statistical trade-off},
	author={Sriperumbudur, Bharath K and Sterge, Nicholas},
	journal={The {A}nnals of {S}tatistics},
	volume={50},
	number={5},
	pages={2713--2736},
	year={2022},
	publisher={{I}nstitute of {M}athematical {S}tatistics}
}

@article{aronszajn1950theory,
	title={Theory of reproducing kernels},
	author={Aronszajn, Nachman},
	journal={Transactions of the {A}merican {M}athematical {S}ociety},
	volume={68},
	number={3},
	pages={337--404},
	year={1950}
}

@article{ullah2018streaming,
	title     = {Streaming Kernel {PCA} with {$\mathcal{O}(\sqrt{n})$} Random Features},
	author    = {Ullah, Enayat and Mianjy, Poorya and Marinov, Teodor Vanislavov and Arora, Raman},
	journal   = {{A}dvances in {N}eural {I}nformation {P}rocessing {S}ystems},
	volume    = {31},
	year      = {2018}
}

@article{fang2024kernel,
	title={Kernel {PCA} for out-of-distribution detection},
	author={Fang, Kun and Tao, Qinghua and Lv, Kexin and He, Mingzhen and Huang, Xiaolin and Yang, Jie},
	journal={{A}dvances in {N}eural {I}nformation {P}rocessing {S}ystems},
	volume={37},
	pages={134317--134344},
	year={2024}
}

@article{rosipal2001kernel,
	title={Kernel {PCA} for feature extraction and de-noising in nonlinear regression},
	author={Rosipal, Roman and Girolami, Mark and Trejo, Leonard J and Cichocki, Andrzej},
	journal={Neural {C}omputing \& {A}pplications},
	volume={10},
	number={3},
	pages={231--243},
	year={2001},
	publisher={Springer}
}

@article{jade2003feature,
	title={Feature extraction and denoising using kernel PCA},
	author={Jade, AM and Srikanth, B and Jayaraman, VK and Kulkarni, BD and Jog, JP and Priya, L},
	journal={Chemical {E}ngineering {S}cience},
	volume={58},
	number={19},
	pages={4441--4448},
	year={2003},
	publisher={Elsevier}
}

@article{teixeira2008kpca,
	title={{KPCA} denoising and the pre-image problem revisited},
	author={Teixeira, Ana R and Tom{\'e}, Ana Maria and Stadlthanner, Kurt and Lang, Elmar Wolfgang},
	journal={Digital {S}ignal {P}rocessing},
	volume={18},
	number={4},
	pages={568--580},
	year={2008},
	publisher={Elsevier}
}

@article{kim2005iterative,
	title={Iterative kernel principal component analysis for image modeling},
	author={Kim, Kwang In and Franz, Matthias O and Sch{\"o}lkopf, Bernhard},
	journal={{IEEE} {T}ransactions on {P}attern {A}nalysis and {M}achine {I}ntelligence},
	volume={27},
	number={9},
	pages={1351--1366},
	year={2005},
	publisher={IEEE}
}

@inproceedings{abrahamsen2009input,
	title={Input space regularization stabilizes pre-images for kernel {PCA} de-noising},
	author={Abrahamsen, Trine Julie and Hansen, Lars Kai},
	booktitle={2009 {IEEE} {I}nternational {W}orkshop on {M}achine {L}earning for {S}ignal {P}rocessing},
	pages={1--6},
	year={2009},
	organization={{IEEE}}
}

@article{bui2019projection,
	title={Projection-free kernel principal component analysis for denoising},
	author={Bui, Anh Tuan and Im, Joon-Ku and Apley, Daniel W and Runger, George C},
	journal={Neurocomputing},
	volume={357},
	pages={163--176},
	year={2019},
	publisher={Elsevier}
}

@article{rasmussen2012nonlinear,
	title={Nonlinear denoising and analysis of neuroimages with kernel principal component analysis and pre-image estimation},
	author={Rasmussen, Peter Mondrup and Abrahamsen, Trine Julie and Madsen, Kristoffer Hougaard and Hansen, Lars Kai},
	journal={Neuro{I}mage},
	volume={60},
	number={3},
	pages={1807--1818},
	year={2012},
	publisher={Elsevier}
}

@article{chang2015critical,
	title={A critical feature extraction by kernel {PCA} in stock trading model},
	author={Chang, Pei-Chann and Wu, Jheng-Long},
	journal={Soft {C}omputing},
	volume={19},
	number={5},
	pages={1393--1408},
	year={2015},
	publisher={Springer}
}

@article{xu2007method,
	title={A method for speeding up feature extraction based on {KPCA}},
	author={Xu, Yong and Zhang, David and Song, Fengxi and Yang, Jing-Yu and Jing, Zhong and Li, Miao},
	journal={Neurocomputing},
	volume={70},
	number={4-6},
	pages={1056--1061},
	year={2007},
	publisher={Elsevier}
}

@incollection{peter20183d,
	title={{3D} face recognition using kernel-based {PCA} approach},
	author={Peter, Marcella and Minoi, Jacey-Lynn and Hipiny, Irwandi Hipni Mohamad},
	booktitle={Computational {S}cience and {T}echnology},
	pages={77--86},
	year={2018},
	publisher={Springer}
}

@article{samuel2016nonlinear,
	title={Nonlinear process fault detection and identification using kernel {PCA} and kernel density estimation},
	author={Samuel, Raphael Tari and Cao, Yi},
	journal={Systems {S}cience \& {C}ontrol {E}ngineering},
	volume={4},
	number={1},
	pages={165--174},
	year={2016},
	publisher={Taylor \& {F}rancis}
}

@article{scholkopf1999input,
	title={Input space versus feature space in kernel-based methods},
	author={Scholk{\"o}pf, Bernhard and Mika, Sebastian and Burges, Chris JC and Knirsch, Philipp and Muller, K-R and Ratsch, Gunnar and Smola, Alexander J},
	journal={{IEEE} {T}ransactions on {N}eural {N}etworks},
	volume={10},
	number={5},
	pages={1000--1017},
	year={1999},
	publisher={IEEE}
}

@article{kwok2004pre,
	title={The pre-image problem in kernel methods},
	author={Kwok, JT-Y and Tsang, IW-H},
	journal={{IEEE} {T}ransactions on {N}eural {N}etworks},
	volume={15},
	number={6},
	pages={1517--1525},
	year={2004},
	publisher={IEEE}
}

@article{honeine2011closed,
	title={A closed-form solution for the pre-image problem in kernel-based machines},
	author={Honeine, Paul and Richard, C{\'e}dric},
	journal={Journal of {S}ignal {P}rocessing {S}ystems},
	volume={65},
	number={3},
	pages={289--299},
	year={2011},
	publisher={Springer}
}

@article{lopez2009svm,
	title={{SVM}-based {CAD} system for early detection of the {A}lzheimer's disease using kernel {PCA} and {LDA}},
	author={L{\'o}pez, MM and Ram{\'\i}rez, J and G{\'o}rriz, JM and {\'A}lvarez, I and Salas-Gonzalez, D and Segovia, F and Chaves, R},
	journal={Neuroscience {L}etters},
	volume={464},
	number={3},
	pages={233--238},
	year={2009},
	publisher={Elsevier}
}

@article{abrahamsen2011regularized,
	title={Regularized pre-image estimation for kernel {PCA} de-noising: input space regularization and sparse reconstruction},
	author={Abrahamsen, Trine Julie and Hansen, Lars Kai},
	journal={Journal of {S}ignal {P}rocessing {S}ystems},
	volume={65},
	number={3},
	pages={403--412},
	year={2011},
	publisher={Springer}
}

@inproceedings{czaja2026comparative,
	title={Comparative Analysis in Pre-image Algorithms of Kernel {PCA}},
	author={Czaja, Wojciech and Ji, Canran},
	booktitle={{T}opology, {A}lgebra, and {G}eometry in {D}ata {S}cience ({TAG-DS} 2025)},
	pages={403--413},
	year={2026},
	organization={{PMLR}}
}

@article{goodfellow2014generative,
	title={Generative adversarial nets},
	author={Goodfellow, Ian J and Pouget-Abadie, Jean and Mirza, Mehdi and Xu, Bing and Warde-Farley, David and Ozair, Sherjil and Courville, Aaron and Bengio, Yoshua},
	journal={{A}dvances in {N}eural {I}nformation {P}rocessing {S}ystems},
	volume={27},
	year={2014}
}

@article{steinwart2001influence,
	title={On the influence of the kernel on the consistency of support vector machines},
	author={Steinwart, Ingo},
	journal={Journal of {M}achine {L}earning {R}esearch},
	volume={2},
	number={Nov},
	pages={67--93},
	year={2001}
}

@article{muandet2017kernel,
	title={Kernel mean embedding of distributions: A review and beyond},
	author={Muandet, Krikamol and Fukumizu, Kenji and Sriperumbudur, Bharath and Scholk{\"o}pf, Bernhard},
	journal={{F}oundations and {T}rends{$\^A$}{\textregistered} in {M}achine {L}earning},
	volume={10},
	number={1-2},
	pages={1--141},
	year={2017},
	publisher={Emerald {P}ublishing {L}imited}
}

@inproceedings{kanagawa2014recovering,
	title={Recovering distributions from {G}aussian {RKHS} embeddings},
	author={Kanagawa, Motonobu and Fukumizu, Kenji},
	booktitle={Artificial {I}ntelligence and {S}tatistics},
	pages={457--465},
	year={2014},
	organization={{PMLR}}
}

@article{rudi2017generalization,
	title={Generalization properties of learning with random features},
	author={Rudi, Alessandro and Rosasco, Lorenzo},
	journal={Advances in {N}eural {I}nformation {P}rocessing {S}ystems},
	volume={30},
	year={2017}
}

@article{micchelli2006universal,
	title={Universal Kernels.},
	author={Micchelli, Charles A and Xu, Yuesheng and Zhang, Haizhang},
	journal={Journal of Machine Learning Research},
	volume={7},
	number={12},
	year={2006}
}

@article{kanagawa2018gaussian,
	title={Gaussian processes and kernel methods: A review on connections and equivalences},
	author={Kanagawa, Motonobu and Hennig, Philipp and Sejdinovic, Dino and Sriperumbudur, Bharath K},
	journal={arXiv preprint arXiv:1807.02582},
	year={2018}
}

@book{wendland2005scattered,
	author    = {Wendland, Holger},
	title     = {{S}cattered {D}ata {A}pproximation},
	series    = {Cambridge Monographs on Applied and Computational Mathematics},
	volume    = {17},
	publisher = {Cambridge University Press},
	address   = {Cambridge},
	year      = {2005},
	isbn      = { 9780521843355}
}

@article{sriperumbudur2011universality,
	title={Universality, Characteristic Kernels and{ RKHS} Embedding of Measures.},
	author={Sriperumbudur, Bharath K and Fukumizu, Kenji and Lanckriet, Gert RG},
	journal={Journal of Machine Learning Research},
	volume={12},
	number={7},
	year={2011}
}

@article{honeine2011preimage,
	title={Preimage problem in kernel-based machine learning},
	author={Honeine, Paul and Richard, Cedric},
	journal={IEEE Signal Processing Magazine},
	volume={28},
	number={2},
	pages={77--88},
	year={2011},
	publisher={IEEE}
}

@inproceedings{honeine2026pre,
	title={A Pre-image Representer Theorem in Machine Learning},
	author={Honeine, Paul},
	booktitle={International Conference on Pattern Recognition},
	pages={401--415},
	year={2026},
	organization={Springer}
}
